\documentclass[letterpaper, 10 pt, journal]{IEEEtran} 
\usepackage{amsmath,amsfonts}
\usepackage{algorithmic}
\usepackage{algorithm}
\usepackage{array}
\usepackage[caption=false,font=normalsize,labelfont=sf,textfont=sf]{subfig}
\usepackage{textcomp}
\usepackage{stfloats}
\usepackage[hyphens]{url}
\usepackage{verbatim}
\usepackage{graphicx}
\usepackage{cite}
\usepackage[colorlinks=true, linkcolor=black, citecolor=black, urlcolor=black]{hyperref}

\usepackage{authblk}

\usepackage{tabularx}
\usepackage{booktabs}
\usepackage{ragged2e}
\newcolumntype{Y}{>{\RaggedRight\arraybackslash}X}

\begin{document}

\title{Foundation Model Driven Robotics: A Comprehensive Review}

\author{Muhammad Tayyab Khan\textsuperscript{†}, Ammar Waheed\textsuperscript{†}, and Seung Ki Moon\textsuperscript{*}
\thanks{This work was supported by the Technology Innovation Program---Industry Technology Alchemist Project (No.~20025702, Development of smart manufacturing multiverse platform based on multisensory fusion avatar and interactive AI) funded by the Ministry of Trade, Industry and Energy (MOTIE, Korea).}
\thanks{Muhammad Tayyab Khan and Seung Ki Moon are with the School of Mechanical and Aerospace Engineering, Nanyang Technological University, Singapore 639798.}
\thanks{Ammar Waheed is with the J. Mike Walker ’66 Department of Mechanical Engineering, Texas A\&M University, College Station, TX 77801, USA.}
\thanks{*Corresponding author: Seung Ki Moon (e-mail: skmoon@ntu.edu.sg).}
\thanks{\textsuperscript{†}Muhammad Tayyab Khan and Ammar Waheed contributed equally to this work.}
}

\markboth{Khan \MakeLowercase{\textit{et al.}}: Foundation Model Driven Robotics: A Comprehensive Review}%
{Khan \MakeLowercase{\textit{et al.}}: Foundation Model Driven Robotics: A Comprehensive Review}

\maketitle

\begin{abstract}
The rapid emergence of foundation models, particularly large language models and vision-language models, has introduced a transformative paradigm in robotics. The models offer powerful capabilities in semantic understanding, high-level reasoning, and cross-modal generalization, enabling significant advances in perception, planning, control, and human-robot interaction. This critical review provides a structured synthesis of recent developments and organizes applications into four directions: simulation-driven robotics, open-world operation, sim-to-real transfer, and adaptable robotic systems. Unlike existing surveys that emphasize isolated capabilities, this work highlights integrated, system-level strategies and evaluates their practical feasibility in real-world environments. Key enabling trends such as procedural scene generation, policy generalization, and multimodal reasoning are discussed alongside core bottlenecks, including limited embodiment, lack of multimodal data, safety risks, and computational constraints. Through this lens, this paper identifies both the architectural strengths and critical limitations of foundation model-based robotics, highlighting open challenges in real-time operation, grounding, resilience, and trust. The review concludes with a roadmap of proposed future research directions for bridging semantic reasoning and physical intelligence through more robust, interpretable, and embodied models.
\end{abstract}


\section{Introduction}\label{sec:introduction}
\IEEEPARstart{T}{he} rapid evolution of large language models (LLMs) and, in general, foundation models has marked a significant milestone in artificial intelligence (AI), particularly in natural language understanding and reasoning. These models, built on transformer architectures \cite{vaswani2017attention} with billions of parameters, are pretrained on massive internet-scale corpora. This equips them with extensive world knowledge and emergent capabilities that go beyond those of smaller models \cite{kim2024survey}. In particular, LLMs such as GPT-3 \cite{floridi2020gpt} have demonstrated impressive few-shot learning abilities without fine-tuning \cite{narang2022pathways, brown2020language}, while more recent models such as GPT-4 \cite{achiam2023gpt, sanderson2023gpt} exhibit advanced reasoning skills and support multimodal capabilities, achieving human-level performance on a variety of benchmarks.
\par
In parallel, rapid advances in robotics, particularly in sensing, learning, control, and planning, have created new opportunities for intelligent physical systems \cite{dorigo2021swarm,karoly2020deep}. Despite this progress, robotic systems still do not have human-level intelligence, especially in terms of the flexibility, adaptability, and generalization required for real-world applications \cite{coombs2020strategic}. They often struggle to transfer knowledge across tasks, adapt to unforeseen scenarios, or exhibit the nuanced decision-making that characterizes human behavior. Traditionally, robot autonomy has been based on explicit programming or narrow task-specific learning \cite{ahn2022can}. These approaches, while effective in constrained settings, limit scalability and present significant challenges in complex, dynamic environments.
\par
In response to these limitations, the recent integration of LLMs into robotics introduces a new paradigm, utilizing their rich semantic knowledge and reasoning abilities to improve communication, planning, and adaptability in robotic agents \cite{kim2024survey}. LLMs can interpret high-level human commands, reason about goals and actions, and even generate low-level control code \cite{guan2023leveraging, liang2023code}. This enables more general-purpose robotic intelligence, allowing robots to tackle a broader range of tasks and environments by drawing on the vast priors learned from language.
\par
However, LLMs alone are agnostic to the physical context. They lack embodiment and do not inherently understand metric quantities, sensor data, or physical dynamics \cite{wang2024large}. The integration of LLMs with robotic systems thus introduces several key challenges. These include integrating language into perception and action, achieving real-time responsiveness, and ensuring safe and reliable behavior. Early research has shown promising results by combining LLMs with vision systems \cite{mu2023embodiedgpt}, feedback mechanisms \cite{liu2024enhancing}, and external knowledge sources \cite{ding2023integrating}. However, the broader challenge of reliably linking language-based intelligence to physical systems in diverse and unpredictable scenarios remains unresolved.
\par
These challenges highlight the need for a comprehensive review of the current state of LLM-driven robotics. Existing key surveys \cite{zeng2023large, kim2024survey, wang2024large, li2025large} focus on traditional subfields such as perception and planning or emphasize specific methods. They tend to overlook the integration of these components in practical settings. A broader perspective is needed, one that systematically examines the relationship between high-level reasoning and low-level control, considers the role of language priors in shaping behavior, and explores the adaptation of general-purpose LLMs to domain-specific constraints.
\par
To address these challenges, this review provides a holistic synthesis of how foundation models and their multimodal extensions are transforming robotics. LLMs and vision-language models (VLMs) are reviewed with a focus on their applications in semantic perception, adaptive planning, goal-directed interaction, and autonomous control. Rather than treating these capabilities in isolation, this review emphasizes integrated strategies that address real-world requirements such as grounding, real-time responsiveness, and safety. Progress in diverse environments, from simulation to the open world, reveals both the potential and the current limitations of LLM-driven robotics. Key bottlenecks, including semantic grounding and real-time performance, are discussed alongside emerging solutions that help bridge the gap between language understanding and physical execution. The review concludes by outlining major trends and open research questions, with the aim of connecting advances in language modeling with the practical demands of embodied intelligence in complex, real-world applications.
\par
The remainder of this paper is organized as follows. Section~\ref{sec:foundation_models} reviews the evolution of LLMs and VLMs and their integration into the core robotic subsystems of perception, planning, control, and human-robot interaction (HRI). Sections~\ref{sec:simulation}--\ref{sec:adaptable} examine four directions in which foundation models are being operationalized at the system level, namely simulation-driven robotics (Section~\ref{sec:simulation}), robots in unstructured, open-world environments (Section~\ref{sec:openworld}), bridging the sim-to-real gap (Section~\ref{sec:sim2real}), and enabling adaptable robotic systems (Section~\ref{sec:adaptable}). Section~\ref{sec:future} discusses the principal limitations of current approaches together with the proposed future research directions. Finally, Section~\ref{sec:conclusion} concludes the paper by summarizing the key findings, the limitations of this review, and future work.
\section{Foundation Models}\label{sec:foundation_models}
The integration of language models into robotic systems is based on a clear understanding of how these models have evolved to support increasingly general, interpretable, and context-aware reasoning. As LLMs become central to enabling high-level cognition in robots, from interpreting user intent to generating executable plans, their architectural and functional progression provides critical context for evaluating their suitability in embodied settings \cite{huang2022inner, huang2022language, song2023llm}. Table~\ref{tab:llm_vlm_summary} provides an overview of several widely adopted models, highlighting their key capabilities and notable applications in robotics. Recent advances in LLMs have been driven by scaling transformer architectures and training on vast and diverse datasets, resulting in emergent capabilities such as instruction following \cite{wei2022emergent, khan2024leveraging}, commonsense inference \cite{brown2020language}, and code generation \cite{gu2023llm}. These capabilities have opened up new avenues in robotics, including natural language planning \cite{gestrin2024nl2plan}, multimodal grounding \cite{allgeuer2024robots}, and interactive behavior \cite{kim2024understanding}.
\par
\begin{figure}
    \centering
    \includegraphics[width=0.9\linewidth]{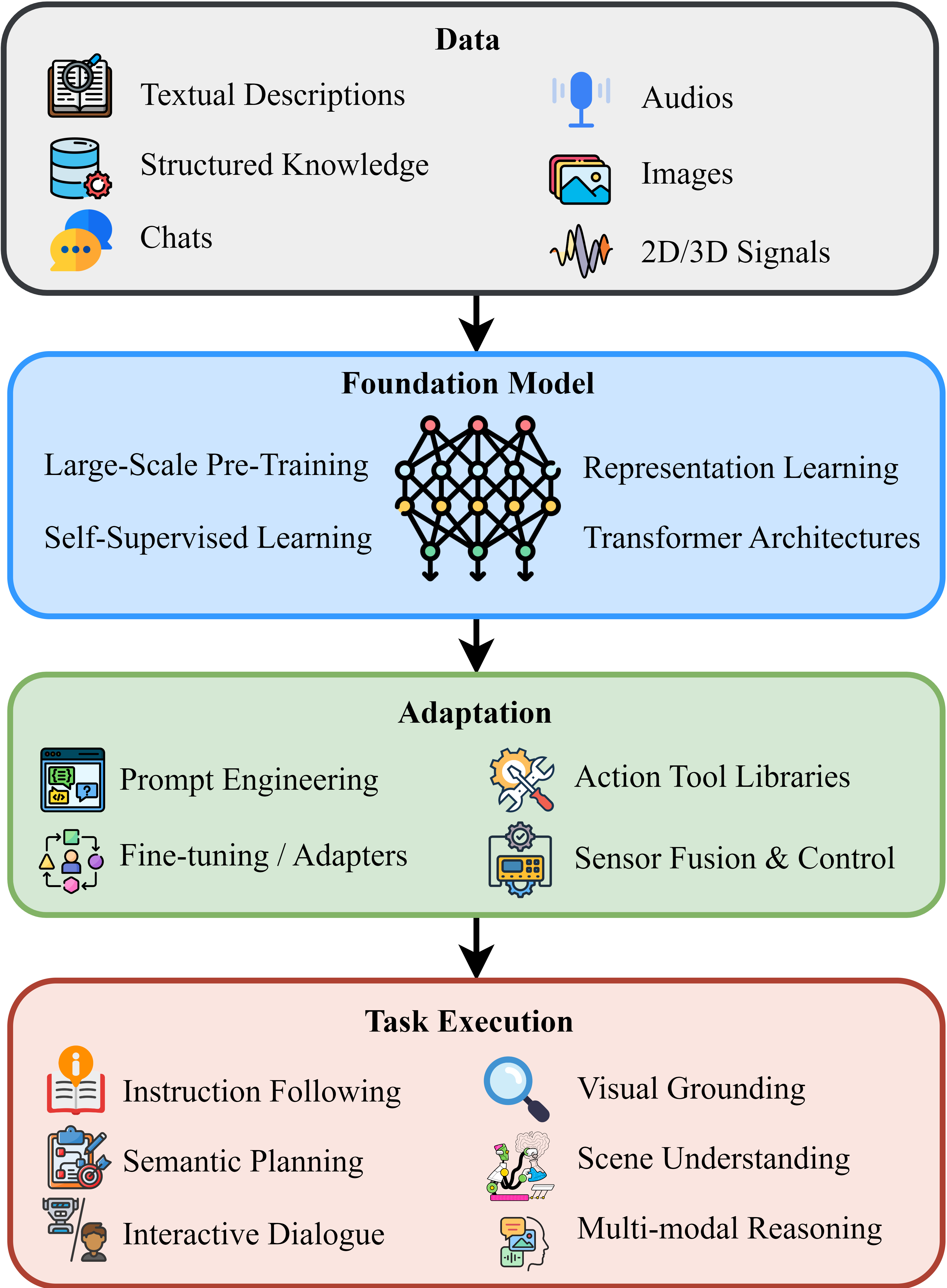}
    \caption{Overview of how foundation models leverage diverse multimodal inputs and scalable pretraining to acquire general-purpose representations, which can then be specialized via lightweight adaptation techniques to support embodied robotic tasks.}
    \label{fig:1}
\end{figure}

The latest generation of models further extends these capabilities through support for visual inputs, refined reasoning, and improved dialogue management. In parallel, open-source alternatives have introduced accessible and fine-tunable models \cite{lykov2024llm, khan2024fine} that are deployable on local hardware, an important consideration for privacy-sensitive real-time robotic systems \cite{li2024personal}. Together, these developments position LLMs as a foundational layer for high-level robot intelligence. Understanding their trajectory from the early milestones to the current state of the art sets the stage for analyzing how such models are being adapted, extended, and embedded within robotic architectures (see Fig.~\ref{fig:1}).
\par
However, robotic systems often rely on rich perceptual input, which requires models that go beyond text processing to interpret visual data in conjunction with language. VLMs extend LLMs with visual perception capabilities, enabling multimodal reasoning for tasks such as image captioning \cite{luu2024questioning}, visual question answering (VQA) \cite{liu2024right}, and referring expression comprehension \cite{hong2024cogvlm2}, all of which are critical for grounded language understanding in robots.
\par
Early efforts introduced joint architectures to align visual and textual features \cite{lu2019vilbert}, laying the foundation for perception-driven interaction \cite{gao2024physically, li2024lmeye, zhao2023chat}. In particular, contrastive learning approaches such as CLIP \cite{radford2021learning} demonstrated open-vocabulary visual recognition using image-text embeddings. This allows robots to perceive objects by name without explicit category training, supporting capabilities like semantic mapping and object retrieval from language prompts. Subsequent VLMs further scaled multimodal training, showing that unified transformer architectures trained on massive image-text corpora could generalize across tasks with minimal supervision. Models such as SimVLM \cite{wang2021simvlm} and Flamingo \cite{alayrac2022flamingo} highlighted how few-shot visual reasoning, captioning, and dialogue about scenes could be achieved without task-specific fine-tuning, offering flexible perception interfaces for embodied agents.
\par
More recent architectures, such as BLIP-2 \cite{li2023blip}, adopt modular strategies that connect frozen image encoders with LLMs using lightweight adapters. This design achieves strong vision performance with reduced training cost, making it practical for robotic deployment. These models enable real-time scene understanding, VQA, and instruction following, all of which are essential for high-level decision-making grounded in the visual environment of a robot. Together, VLMs mark a shift toward generalizable, sample-efficient, and scalable multimodal intelligence, bridging the gap between raw sensory input and language-based cognition in robotic systems.

\begin{table*}[!t]
\centering
 \caption{Overview of Prominent Large Language and Vision-Language Models with Their Relevance to Robotic Applications}
 \renewcommand{\arraystretch}{1.05}
\begin{tabularx}{\textwidth}{>{\RaggedRight\arraybackslash}p{1.8cm} >{\RaggedRight\arraybackslash}p{2.7cm} >{\RaggedRight\arraybackslash}p{3.8cm} Y l}
\toprule
\multicolumn{1}{c}{\textbf{Model}} & 
\multicolumn{1}{c}{\textbf{Architecture}} & 
\multicolumn{1}{c}{\textbf{Key Capabilities}} & 
\multicolumn{1}{c}{\textbf{Applications in Robotics}} & 
\multicolumn{1}{c}{\textbf{Reference}} \\
\midrule
\addlinespace
\multicolumn{5}{c}{\textit{Large Language Models (LLMs)}} \\
\addlinespace
BERT (2018) & Transformer encoder (bidirectional); 110M/340M parameters & Contextual word representations; strong language understanding for natural language processing (NLP) tasks & Parses robotic task steps into structured representations (e.g., behavior trees) for adaptive planning; supports question answering (QA) and dialogue in service robots. & \cite{devlin2019bert, zhou2024llm, silva2024llm} \\
T5 (2019) & Encoder-decoder; up to 11B parameters & Unified text-to-text framework; strong transfer learning capabilities & Enables flexible symbolic task translation, skill sequencing, motion generation, and cross-modal grounding in language-to-action and action-to-language tasks. & \cite{raffel2020exploring, zhang2023large, firoozi2025foundation, caesar2024enabling} \\
Codex (2021) & Decoder (GPT-3-based) & Code generation from natural language instructions & Translates natural language into robot control code for manipulation, path planning, and feedback systems in simulators and real-world tasks. & \cite{chen2021evaluating, liang2023code, mu2024robocodex, shu2024llms} \\
GPT-3.5 (2022) & Decoder (Transformer; GPT-3 lineage); size undisclosed & Instruction-tuned model with high adaptability and reasoning capabilities & Enables real-time path planning, tool retrieval correction, and multi-robot control via natural language commands in simulation and physical setups. & \cite{latif20243p, gao2024integrating, zhao2024applying} \\
InstructGPT (2022) & Decoder (GPT-3 fine-tuned with reinforcement learning from human feedback) & Aligned with human preferences; refined instruction following & Interactive planning and code generation for manipulation and navigation; context-aware object embeddings; perception-planning-action loops via Python (e.g., text-davinci-003 + Segment Anything Model (SAM)/CLIP) & \cite{ouyang2022training, huang2023instruct2act, cong2025overview} \\
LLaMA (2023) & Decoder (Transformer); 7–65B parameters & Efficient, open-source foundation model & Enables object-centric pose prediction, real-time control on edge devices, and multimodal planning in mobile and quadruped robots. & \cite{touvron2023llama, li2024manipllm, liu2024robomamba, sikorski2025deployment, macdonald2024language} \\
\addlinespace
\multicolumn{5}{c}{\textit{Vision-Language Models (VLMs)}} \\
\addlinespace
ViLBERT (2019) & Two-stream vision-language Transformer (BERT-based) & Learns joint visuo-linguistic features; effective for visual QA and reasoning & Powers visual-semantic alignment in robot navigation and embodied QA via object tag anchoring and long-range scene reasoning. & \cite{lu2019vilbert, luo2024transformer} \\
CLIP (2021) & Dual-encoder (vision + text); contrastive learning on 400M pairs & Rich image-text embeddings; strong zero-shot retrieval and classification & Fine-tuned variants (e.g., Robotic-CLIP, CLIP-RT) enable language-conditioned control, visual reward shaping, and object grounding across robots, tasks, and environments including few-shot learning and open-vocabulary perception. & \cite{radford2021learning, kang2024clip, nguyen2024robotic, sontakke2023roboclip, shibata2024clip, shafiullah2022clip} \\
Flamingo (2022) & Multimodal VLM; up to 80B parameters & Few-shot visual reasoning; handles image and text sequences & Supports multimodal manipulation, scene understanding, and reward detection through lightweight adaptation for policy learning and success evaluation. & \cite{alayrac2022flamingo,du2023vision, wang2025roboflamingo, tavassoli2023expanding} \\
BLIP-2 (2023) & Q-Former + frozen vision transformer (ViT) and LLM; two-stage training & Efficient mapping from vision to language; strong zero-shot performance & Enables state recognition of environmental conditions (e.g., open/closed, on/off) via VQA; supports unified perception for mobile robots with no additional training. & \cite{li2023blip, kawaharazuka2024robotic} \\
GPT-4 (2023) & Decoder (Transformer, multimodal variant) & Strong general-purpose reasoning; accepts image inputs & Real-time planning, multimodal feedback, and human-aligned explanations. Gesture generation in humanoids (Alter3), task sequencing with feedback (NUClear), conversational navigation (Dobby), semantic-A* path planning, ethical command rejection, and Robot Operating System 2 (ROS 2) integration. & \cite{achiam2023gpt, yang2023dawn, koubaa2025next, barkley2025semantic, stark2024dobby, o2025exploring, yoshida2023textmotiongroundinggpt4} \\
LLaVA (2023) & CLIP vision encoder + Vicuna (LLaMA-based) LLM; 7B/13B parameters & Multimodal instruction-following; image-language reasoning & Enables context-aware grasp planning, task policy learning, and trajectory evaluation in manipulation tasks. & \cite{liu2023improvedllava, jin2024reasoning, chen2025robo2vlm, liu2024enhancing} \\
PaLM-E (2023) & Decoder-only transformer (PaLM-540B) + ViT-22B; 562B parameters & Embodied reasoning: manipulation planning, multimodal VQA / captioning, zero-shot chain-of-thought (CoT) reasoning, few-shot transfer & Task and motion planning (TAMP), tabletop and mobile manipulation; policy guidance; affordance and failure detection; long-horizon planning. & \cite{driess2023palm, chowdhery2023palm} \\
Gemini (2023) & Decoder-only multimodal Transformer & Direct, no-follow-up predictions; multimodal vision-language understanding & HRI code generation; task-scheduling chatbots; human-robot interaction planning. & \cite{shenawa2025task, sobo2025evaluating, team2023gemini} \\
Claude 3.5 (2024) & Decoder (Transformer, multimodal variant) & High success in spatial reasoning; strong adherence to task constraints in code generation & Demonstrates strong performance in HRI tasks such as coverage path planning and robotic code generation from natural language prompts. & \cite{sobo2025evaluating, kong2024embodied} \\
GR00T N1 (2025) & NVIDIA Eagle-2 VLM and flow-matching diffusion transformer (DiT); 2.2B parameters & Instruction-to-action and multimodal understanding; zero-shot transfer & Bimanual and humanoid manipulation; long-horizon planning; multi-agent coordination. & \cite{bjorck2025gr00t} \\
\bottomrule
\end{tabularx}
\label{tab:llm_vlm_summary}
\end{table*}

\subsection{Integration in Robotics}
Robotic systems (also referred to as agents) are composed of several interdependent subsystems that govern how an agent perceives, plans, acts, and interacts \cite{xie2003fundamentals}. These typically include \textit{Perception} \cite{seminara2019active}, which processes raw sensor data to build an environmental model; \textit{Planning} \cite{galceran2013survey}, which translates high-level goals into sequences of actions; \textit{Control} \cite{kurdila2019dynamics}, which governs low-level actuation and motion execution; and \textit{Human-Robot Interaction (HRI)} \cite{sheridan2016human}, which manages communication and collaboration with users.
\par
\begin{figure}
    \centering
    \includegraphics[width=0.95\linewidth]{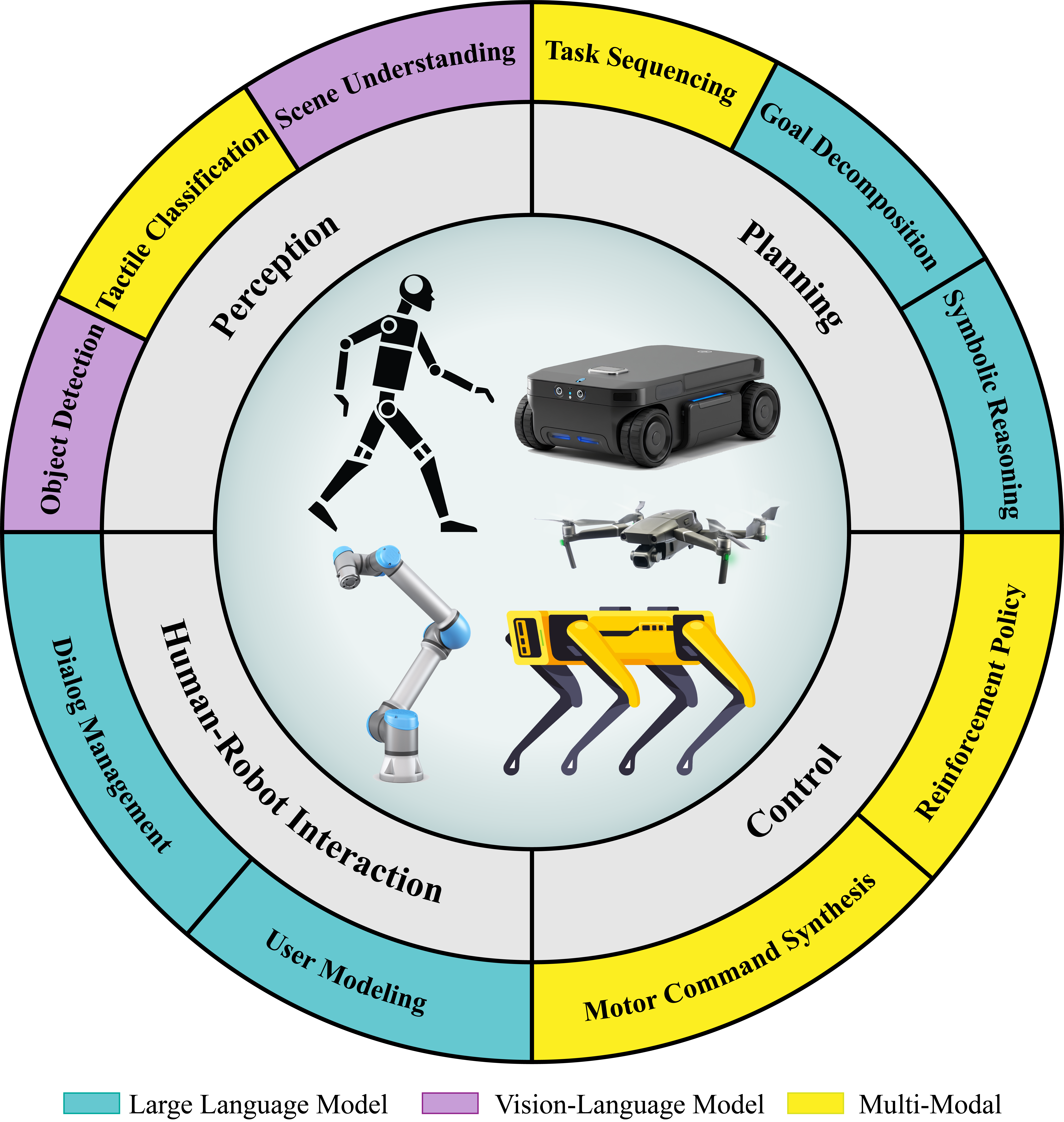}
    \caption{Core components of any robotic system (Perception, Planning, Control, Human-Robot Interaction) with their submodules in the outer ring. Color coding indicates linkage to Large Language Models (LLMs), Vision-Language Models (VLMs), or Multi-Modal Models.}
    \label{fig:2}
\end{figure}

Traditional approaches typically rely on task-specific models for each subsystem, such as convolutional networks for object detection \cite{ren2024deep}, symbolic planners for action sequencing \cite{chen2024language}, and proportional-integral-derivative (PID) (or model-predictive) controllers for motion control \cite{ozen2015practical}. However, these methods often struggle with generalization, adaptability, and semantic grounding, i.e., the ability to link abstract symbols, commands, or language to the perceptual understanding and physical actions of the robot in the real world \cite{sui2025grounding}. These limitations are especially evident when robots operate in dynamic, unstructured environments where predefined models fail to capture all possible edge cases. Fig.~\ref{fig:2} illustrates how foundation models can replace or augment several of these subsystems, enabling more unified and semantically grounded robot behavior. Complementarily, Table~\ref{tab:robot_integration_matrix} summarizes descriptive examples of foundation model integration across these subsystems for humanoids, manipulators, wheeled robots, and quadrupeds.

\subsubsection{Perception}
Visual perception in robotics has seen a leap in generalization due to large-scale VLMs. Traditional robot vision systems were constrained by limited training data and narrowly defined object categories. In contrast, models such as OpenAI’s CLIP learn visual concepts from 400 million image-text pairs, enabling zero-shot recognition of arbitrary objects by name \cite{radford2021learning}. By embedding images and text in a shared space, CLIP lets robots identify new objects or scenes simply by providing a text label, eliminating the need for task-specific training for each category. This capability dramatically improves open-world perception. For example, a robot can be instructed to ``find the spatula'' and use an open-vocabulary detector built on contrastive image-text pretraining (OWL-ViT) to localize it in its camera view, even if it has never seen that exact object before \cite{stone2023open}. Subsequent VLMs such as BLIP-2 further enhanced these capabilities by combining a frozen image encoder with an LLM, achieving strong results in image captioning and VQA with minimal training \cite{li2023blip}. 
\par
Beyond vision, foundation models enable multimodal perception by grounding additional sensor modalities in language semantics \cite{awais2025foundation}. Touch, for instance, is being reimagined as a modality that can integrate with language models. Yang et al. \cite{yang2024binding} introduced UniTouch, which aligns tactile sensor embeddings with pretrained vision-language representations. By mapping touch data to a multimodal semantic space, a robot can perform touch-based tasks in a zero-shot fashion. This is possible because tactile information is grounded in the same feature space as images and words, allowing the robot to predict grasp stability or identify objects by feeling. Recent studies show that language models can integrate commonsense reasoning with tactile data. Yu et al. \cite{yu2024octopi} introduced Octopi, a tactile language model that combines an LLM with a touch sensor to infer object properties. They validated the model in cases where tactile features such as softness relate to abstract concepts such as ripeness through language-based reasoning. This enables sensory input to be interpreted with human-centric semantics, supporting generalization to novel tasks. Foundation models shape robotic perception by enabling open-ended visual recognition, semantic alignment across modalities, and elementary physical reasoning from sensory signals.

\subsubsection{Planning}
LLMs have significantly advanced high-level task planning and decision-making in robotics. Traditional planners required explicit programming or extensive task-specific training for different scenarios \cite{zhou2022review}. In contrast, models such as GPT-3 and GPT-4 provide rich priors on action sequences, commonsense reasoning, and world knowledge. These models can decompose abstract goals into coherent steps without additional training. Huang et al. \cite{huang2022language} demonstrated that a sufficiently large pretrained LLM can convert instructions into a plausible sequence of subtasks in a zero-shot setting. This reflects strong generalization, enabling inference of appropriate action sets for unfamiliar tasks.
\par
However, relying directly on LLM-generated plans can lead to failures when suggested actions exceed the physical capabilities or constraints of the robot. A critical development addressing this limitation is the SayCan framework (PaLM-SayCan) \cite{ahn2022can, chowdhery2023palm}, which integrates the semantic reasoning of an LLM with feasibility estimations based on the robot’s learned affordances. The LLM proposes high-level actions (e.g., “pick up the sponge” or “wipe the spill”), while an affordance model evaluates the likelihood of successful execution in the current context. This architecture enables physical robots to complete complex, multi-step tasks such as “clean up the spilled drink in the kitchen” by grounding abstract instructions in real-world constraints. The combination of general linguistic competence with robot-specific feasibility checking illustrates the role of foundation models in improving semantic grounding. The robot interprets high-level intent and procedural logic from the LLM yet remains constrained to actionable steps through environment-aware grounding.
\par
Another key to adaptable planning is enabling closed-loop reasoning with feedback instead of having the LLM plan once in isolation. The Inner Monologue approach \cite{huang2022inner} exemplifies this by giving the LLM access to environment observations and allowing it to iteratively replan or self-correct through natural language thoughts. If the LLM receives human feedback via success/failure signals or descriptions of what the robot currently sees, it can incorporate this information into an ongoing textual inner dialogue, refining its plan on the fly. This closed-loop, feedback-driven planning dramatically improved success rates on long-horizon tasks compared to one-shot planning.
\par
VLMs are also being used in planning loops for situational awareness. LM-Nav \cite{shah2023lm} uses an LLM to extract landmarks from natural-language route instructions and queries a VLM to ground these landmarks in the robot's visual observations, enabling a robot to follow human directions in outdoor environments. More recently, end-to-end embodied LLMs have appeared. Google’s PaLM-E (562B parameters) \cite{driess2023palm} combines a 540B PaLM \cite{chowdhery2023palm} with a 22B vision transformer \cite{dehghani2023scaling} so that it can plan actions directly from raw images and textual goals. PaLM-E effectively transfers knowledge from internet-scale language and vision data into an embodied agent, allowing it to handle multimodal reasoning tasks. Not only does it address embodied reasoning tasks across multiple robot embodiments, but it also retains general vision-language capabilities (e.g., answering visual questions) within the same model. This kind of holistic model points to future robots that treat planning as an extension of dialogue and perception, richly grounding decisions in what the robot sees and knows. In summary, foundation models have made robot planning more general and flexible. Robots can now leverage LLMs and VLMs to break down novel tasks, use knowledge beyond their direct experience, and adjust plans in real time, all of which are crucial for open-world autonomy.
\begin{table*}[!t]
\centering
\caption{Descriptive Examples of Foundation Model Integration Across Robotic Subsystems and Robot Types}
\renewcommand{\arraystretch}{1.25}
\begin{tabularx}{\textwidth}{l Y Y Y Y}
\toprule
\multicolumn{1}{c}{\textbf{Subsystem}} & 
\multicolumn{1}{c}{\textbf{Humanoids}} & 
\multicolumn{1}{c}{\textbf{Manipulators}} & 
\multicolumn{1}{c}{\textbf{Wheeled Robots}} & 
\multicolumn{1}{c}{\textbf{Quadrupeds}} \\
\midrule

\textbf{Perception} 
& Scene-aware object disambiguation from dialogue (e.g., “pick up the cup to the left of the book”) 
& Multi-view affordance detection using open-vocabulary vision models 
& Semantic navigation via open-set object recognition 
& Terrain-aware object detection for unstructured or outdoor environments \\

\textbf{Planning} 
& Task decomposition for high-level instructions (e.g., “set the table for dinner”) 
& Assembly planning using LLM-generated action sequences and tool affordances 
& Navigation plan synthesis from natural language goals (e.g., “go to the kitchen”) 
& Path generation incorporating terrain constraints and verbal mission goals \\

\textbf{Control} 
& Adaptive motion generation from high-level prompts (e.g., “walk over and wave”) 
& Low-level control code synthesis for stacking, sorting, or manipulating parts 
& Dynamic obstacle avoidance based on text-guided policy modulation 
& Gait and balance adaptation via LLM-suggested behavior strategies \\

\textbf{Interaction} 
& Multimodal dialogue including gesture, gaze, and context-rich conversation 
& Instruction refinement through natural language correction (e.g., “place it upright”) 
& Conversational feedback during navigation (e.g., “turn after the red sign”) 
& Remote mission updates via verbal feedback or contextual corrections \\

\bottomrule
\end{tabularx}
\label{tab:robot_integration_matrix}
\end{table*}


\subsubsection{Control}
At the level of motion control and policy execution, foundation models introduce mechanisms for generalization and adaptability in robotics. One research direction uses large models as policy networks directly, under the idea of a generalist controller that can operate across many tasks and embodiments. A notable example is DeepMind’s Gato \cite{reed2022generalist}, a transformer model trained on data from more than 600 tasks, including gameplay, image captioning, and physical robot manipulation. Gato can produce text or torque commands depending on the input modality, demonstrating cross-domain policy flexibility. Although not superior to specialist models, it validates the feasibility of multi-task multimodal control architectures. 
\par
Building on this concept, the Robotics Transformer series introduced scalable policy learning. RT-1 \cite{brohan2022rt} and RT-2 \cite{brohan2023rt}, developed by Google, used extensive datasets and pretrained visual and language representations to map visual inputs to robotic actions. RT-2 fine-tunes a pretrained VLM on robot data, thereby injecting web-scale semantic knowledge into a robot’s control policy \cite{zeng2023large}. This allows the model to execute commands based on high-level concepts, such as identifying and manipulating an object described in natural language even if it was not seen during training. Such capabilities represent progress toward open-world control policies that generalize beyond the training distribution.
\par
Another interesting development is the use of LLMs to generate code for robot policies on the fly \cite{burns2024genchip, ji2025genswarm}. Instead of training a monolithic network to output motor commands, the model synthesizes scripts that invoke motion primitives and perception routines. Liang et al. \cite{liang2023code} demonstrated this concept with the Code as Policies framework. Given a natural language command, it returns a script implementing the desired behavior using predefined robot application programming interfaces (APIs). The generated code supports logic, loops, and mathematical operations, enabling dynamic policy synthesis. For example, if instructed to “draw a circle around the cup,” the LLM could generate a small program that queries the position of the cup from vision, then computes waypoints in a circle and sends those to the robot controller. This approach showed impressive generalization with a suitable API, which allowed the same LLM to write policies for a variety of tasks (navigation, manipulation, etc.) without task-specific training, handling novel instructions by stitching together known functions in new ways. Code generation policies also provide a form of interpretable robot reasoning: the resulting code can be inspected or adjusted by humans, aiding transparency and debugging. Executing arbitrary code raises safety considerations, but if constrained to vetted libraries and simulators, it offers a powerful tool for open-world operation. 
\par
Between large pretrained policy networks and on-the-fly code generation, control-level foundation models are enabling robots to respond to scenarios that were never explicitly seen in training. This is achieved either through generalization of policy representations or by synthesizing new ones on demand. Early results show that robots can solve problems that require spatial reasoning or novel sequences of motor actions through LLM-generated code that standard end-to-end controllers would struggle with \cite{liang2023code}. As these techniques mature, robot control is expected to become more broadly generalizable and semantically aware, closing the gap between high-level intent and low-level execution.

\subsubsection{Human-Robot Interaction}
Foundation models are reshaping HRI by enabling robots to interpret natural language and visual input with contextual fluency. LLMs can convert conversational instructions into executable policies, reducing the need for technical expertise. For example, Vemprala et al. \cite{vemprala2024chatgpt} showed that ChatGPT can translate user commands into robot control code while also supporting iterative feedback through dialogue. This allows non-expert users to instruct and adjust robot behavior using natural language, while the robot can respond, explain its actions, or request clarification as needed.
\par
VLMs such as PaLM-E and BLIP-2 further enrich the interaction by grounding the dialogue in real-world perception. These models allow robots to describe and reason about their surroundings, enhancing transparency and enabling explanation of actions. For example, a robot can respond to queries or infer corrective actions based on visual context \cite{nasiriany2024pivot}. Beyond perception and control, foundation models embed commonsense and social knowledge that supports more intuitive behavior. They can infer human preferences, anticipate collaborator actions, and conform to social norms, such as handing over tools appropriately or modulating tone in context-sensitive ways \cite{zhang2023large}. Moreover, embodied LLM frameworks that couple GPT-4 reasoning with retrieval-augmented knowledge and visual and force feedback have completed long-horizon, user-requested tasks in unpredictable environments \cite{mon2025embodied}.
\par
It should be noted that these advances also bring new challenges. Language models can be prone to errors and hallucinations if not properly constrained \cite{xu2024hallucination}, and their substantial computational requirements pose integration hurdles for real-time robotic control \cite{jeong2024survey}. Ensuring safety and reliability when an LLM is writing robot code or when misclassifications of a vision model could lead to poor grasps remains an active area of research \cite{wu2024highlighting}. Techniques such as grounding through value functions \cite{cohen2024survey}, closed-loop feedback \cite{rani2006affective}, and human-in-the-loop oversight \cite{skubis2023humanoid} are being explored to mitigate these issues. Meanwhile, research is pushing into even more modalities. Recent work on audio language models \cite{wu2024towards} and cross-modal transfer \cite{xu2024cross} suggests that future foundation models may unify sight, sound, touch, and language, giving robots an even more human-like understanding of context. In the coming years, more complex and integrated “robot foundation models” are expected to drive further improvements in general performance. Despite the challenges, foundation models are fundamentally transforming how robots are programmed and how they behave, moving closer to robots that can truly comprehend and operate in complex, unscripted environments.

\subsection{Toward Holistic, Resilient, and Scalable Robotics}\label{sec:holistic}
The preceding review of LLMs and VLMs in perception, planning, control, and HRI highlights both remarkable progress and persistent gaps. Foundation models have enabled unprecedented capabilities, including open-vocabulary visual perception, high-level task reasoning, and code generation for robot behaviors, indicating their potential for zero-shot generalization in robotics \cite{firoozi2025foundation}. However, several key limitations remain evident across these domains. First, today’s foundation models still struggle with real-time onboard operation due to slow inference and vast computational requirements, which conflict with the speed and resource constraints of robotic systems \cite{jeong2024survey}. Second, there is a lack of robot-specific training data and experience; models mostly pretrained on internet-scale text or image data often lack the physical grounding and long-tail robotic corner cases needed for reliable deployment \cite{hu2023toward}. This data gap hinders robust perception and planning in diverse settings, since purely web-scale training does not guarantee coverage of safety-critical scenarios or low-level control nuances. Third, current approaches are largely limited in modality and scope, where many LLM-based policies rely solely on language or vision and cannot yet ingest the full spectrum of real-world inputs (e.g., audio, tactile, or proprioceptive data) \cite{jones2025beyond}. This limits situational awareness and the ability to respond to dynamic physical interactions. Finally, safety and ethical challenges pose barriers that extend beyond technical performance. LLM-driven robots can produce actions or instructions that are biased or unsafe, especially if prompt attacks or distribution shifts exploit their lack of rigorous validation \cite{ravichandran2025safety}. Ensuring trustworthy behavior under uncertainty remains an open problem. Despite substantial progress, today’s foundation model-based robots are still not fully holistic. They do not yet seamlessly integrate all necessary modalities and knowledge, nor are they sufficiently resilient to novel conditions or scalable in terms of data and learning efficiency. These shortcomings highlight the need for more comprehensive and robust frameworks, and researchers are already exploring new directions to advance foundation models further. Sections~\ref{sec:simulation}--\ref{sec:adaptable} discuss four such promising directions that address these critical gaps.

\section{Simulation-Driven Robotics}\label{sec:simulation}
Foundation models are reshaping robot simulation frameworks by automating environment synthesis, procedural content creation, simulation scripting, and multi-robot coordination. As simulators such as NVIDIA Isaac Sim, Unreal Engine 5 (UE5), Unity, and Gazebo advance in realism, manually creating diverse and meaningful training scenarios is becoming increasingly labor-intensive. Recent studies utilize foundation models to construct rich simulation scenarios for various types of robots and tasks with minimal human intervention, mitigating data scarcity and improving generalization \cite{katara2024gen2sim, lin2025proc4gem}.

\subsection{Procedural Environment and Task Generation}
\begin{figure}
    \centering
    \includegraphics[width=0.9\linewidth]{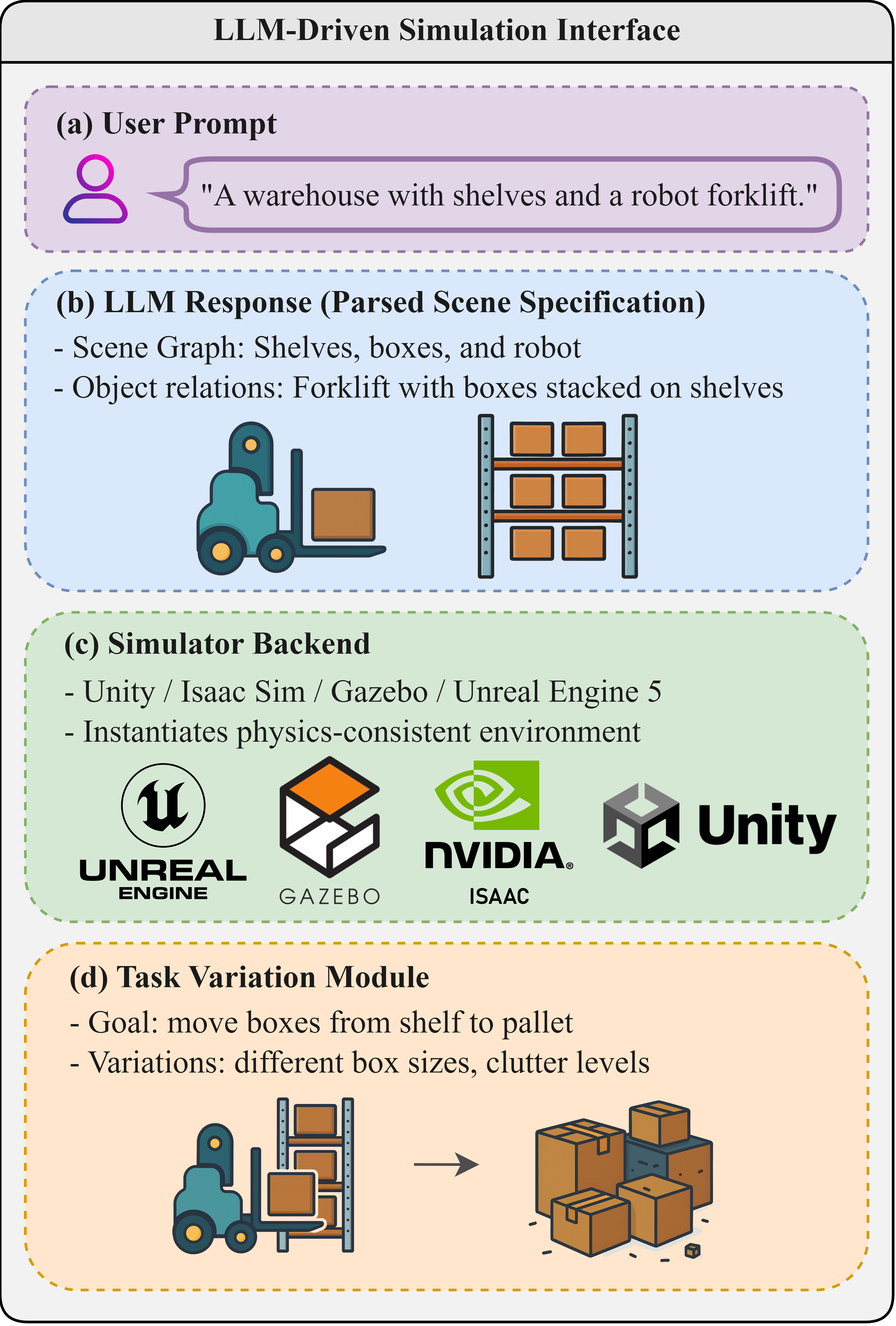}
    \caption{Overview of the LLM-driven simulation interface. (a) The system begins with a natural language prompt from the user describing a target environment. (b) The LLM generates a structured scene specification, capturing relevant objects and their spatial relationships. (c) A simulator backend instantiates a physics-consistent virtual environment based on this specification. (d) The task variation module introduces goals and environmental diversity to enable robust task execution and learning.}
    \label{fig:3}
\end{figure}

Foundation models are increasingly applied to the generation of procedural environments by directly translating natural language prompts into simulation assets, scene layouts, and task specifications \cite{katara2024gen2sim}. Instead of manual world file creation, language models generate complete scene configurations, including object placement, semantic relations, and environmental conditions. Fig.~\ref{fig:3} illustrates a representative LLM-driven simulation interface, highlighting how a language prompt is transformed into a structured scene graph, instantiated in a simulator backend, and diversified through task variation modules. This approach enables rapid scaling of training data in diverse manipulation and navigation scenarios.
\par
Several methods leverage natural language to generate simulator-specific world files across multiple simulation engines by parsing scene descriptions and dynamically retrieving objects from public 3D asset libraries such as Objaverse and Gazebo Fuel \cite{yang2024holodeck, afzal2021gzscenic, karavaev_worldcreator}. These methods use semantic reasoning to place objects in physically plausible arrangements, improving diversity over manually curated scenes. However, challenges remain in spatial reasoning accuracy, object collisions, and unrealistic configurations due to limited physical understanding, leading to hallucinated or unstable environments.
\par
Beyond static environment generation, foundation models enable semantic domain randomization by procedurally varying task parameters while preserving underlying task coherence. Scene graphs and structured semantic representations are automatically synthesized to define object relationships, spatial dependencies, and task goals, allowing controlled manipulation of object types, positions, and relational constraints \cite{gao2025genmanip}. Such structured variation improves dataset diversity and supports better generalization to unseen scenarios, though errors in relational grounding may introduce logically inconsistent task setups when not properly constrained.

\subsection{Code-Driven Simulation Design}
Foundation models increasingly act as autonomous code generators that translate high-level textual task descriptions into executable simulation scripts \cite{xu2023creative}. These scripts specify environment geometries, object attributes, task objectives, reward functions, and simulator configurations using programming languages and description formats such as Python, C\#, Unified Robot Description Format (URDF), and Universal Scene Description (USD). Fine-tuning code-specialized models on synthetic datasets generated by large models further improves the robustness and adaptability of code to novel tasks \cite{wang2023gensim}.
\par
Beyond task description parsing, foundation models generate end-to-end simulation pipelines by producing structured control code that integrates perception modules, motion planners, and manipulation skills \cite{chen2024roboscript}. Chain-of-thought (CoT) prompting is often used to scaffold model reasoning, improve correctness, and allow for error traceability during simulation validation stages. Although these methods streamline the development of complex robotics pipelines, they remain sensitive to ambiguous task specifications and can propagate incorrect reasoning into unsafe control policies \cite{zawalski2024robotic}. Simulation platforms with programmatic interfaces, such as AI2-THOR \cite{kolve2017ai2}, NVIDIA Isaac Sim, and Omniverse, are increasingly compatible with such code generation workflows. Although native LLM integration remains limited, emerging prototypes enable ChatGPT-like interfaces to generate Omniverse Replicator scripts for synthetic dataset generation and randomized scene creation. Similarly, UE5's Procedural Content Generation (PCG) framework offers integration potential, allowing foundation models to dynamically author procedural rules for terrain generation, dynamic obstacles, and asset placement, opening new frontiers for simulation diversity.
\par
Recent extensions introduce intermediate domain-specific languages that bridge natural language input with executable low-level control programs for aerial robotics. For example, TypeFly has the LLM emit compact programs in a domain-specific language (MiniSpec) that a lightweight runtime executes, reducing response time by up to 62\% compared with Python-based plans \cite{chen2023typefly}. Although this approach offers lower computational overhead and faster response times, it presumes accurate and complete translation from language to program, making system performance vulnerable to ambiguity, underspecification, and context drift.

\subsection{Multi-Robot and Multimodal Simulation Tasks}
\begin{figure}
    \centering
    \includegraphics[width=1\linewidth]{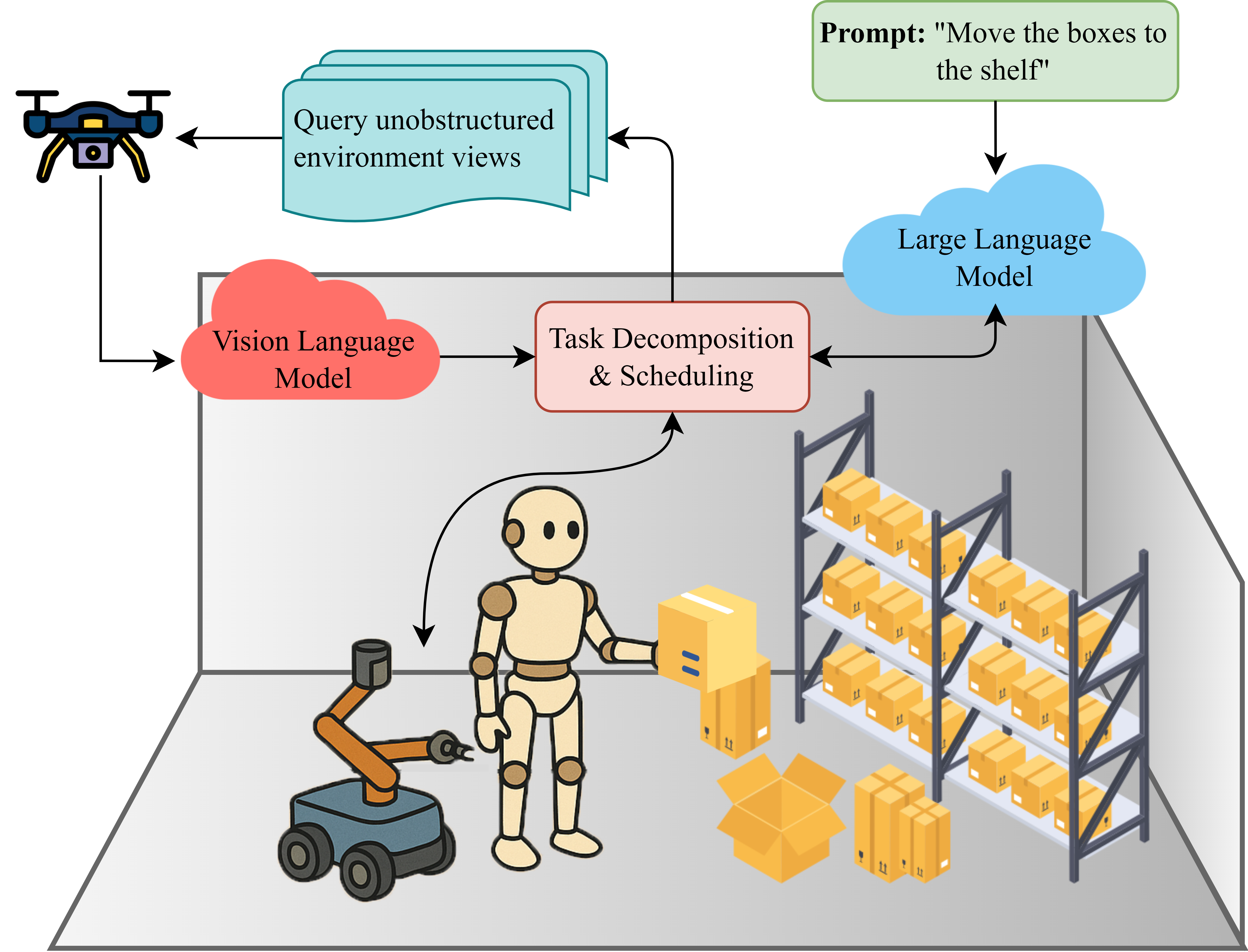}
    \caption{Multi-robot coordination simulation driven by large language and vision-language models. The high-level instruction is decomposed into subtasks and scheduled for a drone, mobile manipulator, and humanoid robot. The drone provides dynamic views to a VLM for real-time scene understanding, enabling robust task execution and validation.}
    \label{fig:4}
\end{figure}

Foundation models also support multi-robot coordination by decomposing complex instructions into structured subtasks, forming robot coalitions, and allocating subtasks for ordered and parallel execution \cite{kannan2024smart}. Such approaches are typically benchmarked in simulation (e.g., AI2-THOR) before being demonstrated on physical robots, allowing coordination errors to be identified before physical execution. Fig.~\ref{fig:4} illustrates a representative system in which high-level instructions are processed by LLMs and VLMs to schedule subtasks across a drone, mobile manipulator, and humanoid robot, enabling coordinated execution within a shared environment. Although these methods improve scalability across heterogeneous agent teams, task allocation may fail under conflicting objectives, resource constraints, or unexpected inter-agent dependencies. Recent work also integrates locally hosted language and vision-language models with PX4 flight control and ROS 2 middleware to enable natural language control of unmanned aerial vehicle (UAV) agents \cite{lim2025taking}. Such systems minimize cloud dependency and have been validated both in simulation and on a physical quadcopter.
\par
Multimodal simulation pipelines combine LLMs with VLMs to create interactive environments. GenManip \cite{gao2025genmanip} uses an LLM to generate scene-graph-based manipulation tasks in simulation and benchmarks modular systems that combine foundation models for perception, reasoning, and planning against end-to-end policies, finding that the modular systems generalize better to novel instructions and object configurations. Beyond configuring simulations, foundation models also benefit from synthetic data generated in simulation environments. Researchers have used Omniverse to procedurally synthesize spatial reasoning datasets, such as camera object pose datasets, to fine-tune VLMs for spatial understanding tasks \cite{nvidia_physicalai_spatial_warehouse}. These self-reinforcing training pipelines, in which simulation improves foundation models that in turn better configure simulations, are emerging as powerful mechanisms for iterative model enhancement.

\section{Robots in Unstructured, Open-World Environments}\label{sec:openworld}
Recent advances in foundation models have shown remarkable potential to expand robotic autonomy beyond narrowly scripted tasks into open-world, unstructured environments \cite{naderi2024foundation}. Unlike classical control pipelines, these systems leverage multimodal embeddings, semantic reasoning, generative policy learning, and predictive forecasting to generalize across objects, tasks, and settings not seen during training \cite{szot2023large}. This section discusses the principal methods that operationalize foundation models for such open-world robotic systems, emphasizing their architectural design, operational capabilities, and emerging limitations.
\subsection{Multimodal Reasoning and Hierarchical Task Decomposition}
To address the complexity of open-world environments, several methods integrate multimodal perception with language-conditioned reasoning to generate adaptive task plans. A prominent strategy involves decomposing high-level commands into chain-of-action sequences that map abstract goals to executable locomotion and manipulation primitives \cite{hao2025embodied}. VLMs first parse the scene and instructions to extract semantic and spatial information, followed by LLMs that synthesize temporally ordered action chains. These decompositions incorporate affordance reasoning, ensuring that the generated subgoals are physically grounded in the robot’s embodiment and current environmental context. This affordance-driven breakdown helps mitigate infeasible or unsafe subactions by continuously checking the availability, graspability, and reachable surfaces of objects.
\par
Extensions of this approach incorporate embodiment-specific simulation loops that refine coarse plans through feasibility checks. For example, QuadrupedGPT \cite{mei2024quadrupedgpt} combines language-based subgoal proposals with a Location-Simulation-Selection (LSS) strategy for terrain-adaptive locomotion, in which the multimodal foundation model first estimates coarse ranges for gait parameters (e.g., body height, stepping frequency, and foot swing height), candidate parameter sets are then evaluated in physics simulation, and the best-performing set is selected for execution. By combining high-level semantic reasoning with physics-informed low-level controllers, these systems enable robust adaptation to variable terrains while maintaining stability across long-horizon tasks. However, these systems remain reliant on accurate environmental perception; sensor noise, misclassification, or incomplete semantic maps can still propagate downstream, leading to execution failures if not sufficiently filtered.
\par
An alternative perspective focuses on embodied orchestration frameworks that leverage multimodal foundation models not for direct control but for scalable autonomous task generation. Systems such as AutoRT \cite{ahn2024autort} use a VLM to describe the robot scene in natural language, which is then used to prompt language models to generate novel task proposals. To ensure safety and physical feasibility, these generated instructions are screened by an LLM-based filter guided by a robot constitution, a set of foundational, safety, and embodiment rules that rejects unsafe, impractical, or embodiment-incompatible tasks. This approach enables continuous self-directed exploration and dataset generation in unstructured real-world environments, producing highly diverse experience buffers for downstream learning. However, while constitution-based filtering substantially reduces unsafe proposals, edge-case failures and unsafe recommendations remain a residual concern, often necessitating limited human oversight.
\subsection{Generalist Policy Models and Cross-Embodiment Control}
\begin{figure}
    \centering
    \includegraphics[width=1\linewidth]{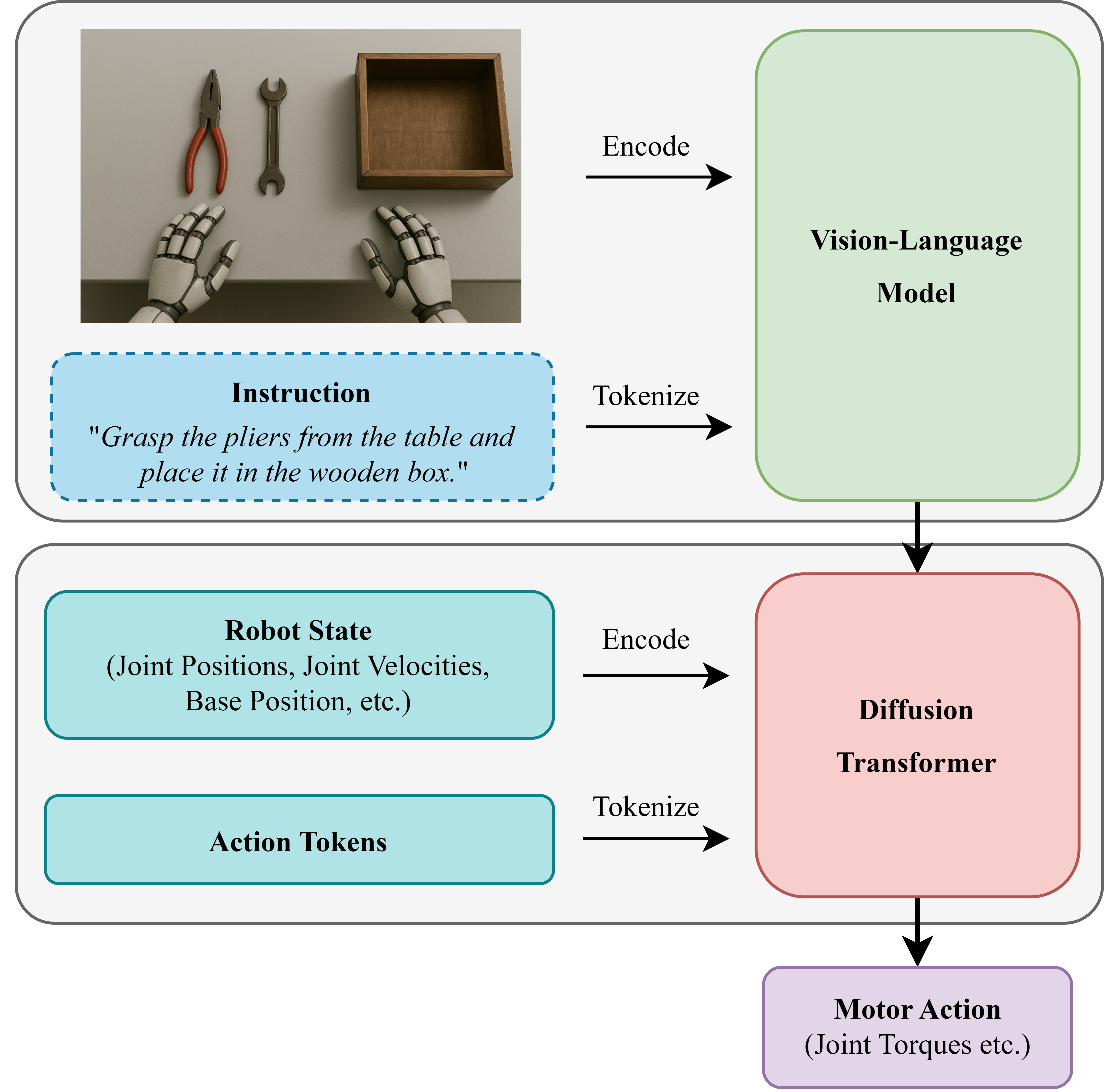}
    \caption{A typical Vision-Language-Action (VLA) foundation model architecture for synthesizing a humanoid's motor torque commands.}
    \label{fig:5}
\end{figure}

Beyond high-level planning, open-world operation requires adaptable control policies that can be generalized across tasks, environments, and embodiments. Generalist policy architectures address this through large-scale vision-language-action (VLA) models that learn end-to-end mappings from sensory input and task descriptions to continuous motor outputs. Fig.~\ref{fig:5} shows a representative VLA architecture, where a VLM encodes visual observations and instructions, and a diffusion transformer maps this multimodal input along with the state of the robot to executable motor actions. NVIDIA's GR00T N1 \cite{bjorck2025gr00t} exemplifies this class of models through a similar architecture design, where a vision-language transformer interprets observations and goals, while a diffusion-transformer policy generates temporally coherent high-dimensional motor commands. The generative nature of diffusion policies enables the production of smooth, flexible trajectories that adapt to task variations in real time.

\par
Training such models involves heterogeneous data comprising real-world demonstrations, synthetic simulations, and large-scale video corpora that capture human motion. This diverse exposure allows a single policy to generalize across multiple robot embodiments, including humanoids and manipulators, while requiring limited fine-tuning for new platforms. However, the training process is computationally intensive, and maintaining stability during diffusion policy optimization presents additional challenges. Moreover, despite their impressive generalization, safety remains a persistent concern in real-world deployment, particularly when ambiguities in perception or instructions lead to unintended behaviors that require constraint-based safeguards.
\par
Complementary to policy models, world models offer predictive forecasting capabilities that inform long-term planning. Humanoid World Models (HWM) by Ali et al. \cite{ali2025humanoid} demonstrate lightweight video prediction architectures that forecast future egocentric observations conditioned on candidate action sequences. Two generative formulations, namely masked transformers and flow-matching models, are trained on humanoid demonstrations to forecast future video frames, allowing efficient simulation of future scene evolution without requiring full 3D state estimation or high-fidelity physics engines. This enables robots to anticipate plausible outcomes, such as potential collisions or object displacements, when evaluating action sequences. Despite their compact size, these models remain limited in representing explicit object-level states, which can result in uncertainty when encountering unfamiliar or ambiguous scenes.

\section{Bridging the Sim-to-Real Gap}\label{sec:sim2real}
Bridging the simulation-to-reality (Sim2Real) gap remains challenging due to unmodeled dynamics and sensor discrepancies \cite{chukwurah2024sim}. Foundation models offer promising solutions by facilitating reliable transfer through domain-invariant representations and improved policy generalization \cite{yu2024natural}. Emerging methods leverage LLMs not as direct task controllers, but rather to refine simulations or training processes, resulting in robust real-world performance. Such approaches exploit the extensive prior knowledge and abstraction capabilities inherent in foundation models, enabling efficient transfer across robotic platforms. Recent surveys emphasize the potential of foundation model-based methods in addressing longstanding Sim2Real challenges \cite{da2025survey}.

\subsection{Semantic Alignment and Multimodal Representations}
Foundation models increasingly facilitate consistent semantic representations across simulation and reality by leveraging high-level features such as object identities, spatial relations, and task context as common semantic anchors \cite{noorani2025abstraction}. Yu et al. \cite{yu2024natural} utilized natural language descriptions to guide an image encoder toward learning task-relevant, domain-invariant visual features while ignoring domain-specific details such as texture or lighting. In object manipulation tasks, where policies were trained on abundant simulated data and only a handful of real-world demonstrations, this language-based pretraining approach outperformed established pretraining baselines such as CLIP and R3M \cite{nair2022r3m} by 25–40\%. This result reinforces the effectiveness of language as a robust cross-domain descriptor, a perspective also supported by the TIAMAT program of the U.S. Defense Advanced Research Projects Agency (DARPA), which advocates semantic anchors \cite{noorani2025abstraction}.
\par
Similarly, VLMs have been employed for perception-level Sim2Real transfer. Fine-tuning small VLMs on simulated data to generate detailed physical scene descriptions, which then serve as context for larger VLMs, improves physical reasoning and transfers from simulation to real scenes \cite{balazadeh2024synthetic}. Liu et al. \cite{liu2025fetchbot} showed that using depth maps predicted by a foundation model, which emphasize geometry and ignore texture details, significantly improved transfer robustness in cluttered environments.

\subsection{Generative Simulation and Domain Adaptation}
Generative models, ranging from generative adversarial networks (GANs) to diffusion-based foundation models and pretrained simulators, enhance Sim2Real transfer by generating realistic and diverse training data, thus improving domain randomization and adaptation \cite{zhang2025generative}. Traditional randomization methods vary physics or rendering parameters to generalize policies \cite{chen2021understanding, muratore2018domain}. In contrast, generative approaches offer data-driven and high-fidelity augmentations. For example, Zhao et al. \cite{zhao2024exploring} found that, when conditioned on simulator-generated labels, diffusion-based synthesis (ControlNet) produced driving images with fewer artifacts and greater structural fidelity than GAN-based methods. This suggests that modern diffusion models can serve as effective tools for style transfer and data augmentation in Sim2Real, particularly in preserving critical scene structure.
\par
Extending beyond static images, generative models now dynamically create complex 3D environments. Yu et al. \cite{yu2025adept} proposed ADEPT, an adaptive diffusion technique generating increasingly challenging off-road terrains informed by policy performance. ADEPT continuously introduces novel simulation scenarios, yielding policies that outperform those trained on procedurally generated or natural terrains in simulation and that transfer zero-shot to real-world off-road navigation. Such use of foundation models essentially automates an open-ended curriculum of domain randomization, a significantly more scalable and diverse augmentation process than manual parameter tuning.
\par
It is worth noting that foundation models also play a role in direct visual domain adaptation techniques. Instead of retraining policies from scratch on real data, adaptation methods seek to adjust the simulator outputs or learned features to resemble real ones. Classic approaches include latent feature alignment or GAN-based image-to-image translation \cite{jang2024bridging}. Pretrained visual models offer a complementary route: self-supervised vision transformers and robust representations such as R3M \cite{nair2022r3m} or MVP \cite{radosavovic2023real} provide inherent invariance that directly benefits Sim2Real transfer \cite{biruduganti2025bridging, chen2024sugar}. Recent methods extend this strategy using multimodal foundation models like CLIP, substantially simplifying policy learning \cite{kang2024clip}. Collectively, generative and representation-focused foundation models enrich simulation by producing high-quality data and robust feature spaces, thereby reducing reliance on extensive real-world training.

\subsection{Policy Learning and Transfer with Large Models}
Foundation models increasingly shape policy learning directly, enhancing Sim2Real transfer by automating simulator design, reward definition, and training curricula. For example, DrEureka \cite{ma2024dreureka} leverages LLMs to propose reward terms and domain randomization parameters, replacing manual engineering with automated and knowledge-driven configuration. This approach achieved Sim2Real performance comparable or superior to that of human-designed configurations, notably solving a novel task, quadruped balancing and walking atop a yoga ball, without iterative manual design.
\par
Foundation models also guide exploration and improve sample efficiency. Chen et al. \cite{chen2024rlingua} demonstrated using an LLM-generated preliminary policy to bootstrap reinforcement learning (RL) agents. In effect, the language model’s knowledge is converted into an exploratory policy that helps the RL agent avoid catastrophic moves early on and focus on promising behaviors, which reduces the number of simulator interactions required and accelerates convergence to a robust policy. Likewise, as reviewed by Da et al. \cite{da2025survey}, foundation models have been used to improve dynamics modeling for Sim2Real, where LLM-derived knowledge of system dynamics yields more accurate forward predictions and improved real-world policy performance. Additionally, Jiao et al. \cite{jiao2023swarm} employed an LLM (GPT-3.5) for drone swarm choreography, translating textual instructions and music features into waypoints that a safe swarm motion planner refines into collision-free, dynamically feasible quadcopter trajectories. These methods illustrate foundation models' role as interactive training loop components rather than passive tools.
\par
However, foundation model applications face critical challenges, notably hallucinations, computational costs, and evaluation complexity. Hallucinated outputs from LLMs or vision models can produce unsafe or invalid actions, necessitating grounding strategies such as retrieval augmentation or safety validation filters, although retrieval augmentation alone does not guarantee safer outputs \cite{an2025rag}. Computational efficiency remains a key concern, motivating the use of model distillation techniques to preserve the advantages of foundation models while reducing computational overhead. Additionally, evaluating the reliability of these learned systems through conventional parameter-based methods (e.g., mean squared error) is often insufficient, as they do not capture critical qualitative differences in dynamics or control behaviors. Instead, model-based validation methods offer a deeper understanding. Approaches such as control-theoretic metrics that quantify Sim2Real dynamics divergence \cite{waheed2025quantifying} explicitly expose discrepancies in unmodeled physical interactions, including friction and other subtle dynamic effects. These discrepancies often have a direct impact on real-world performance.
\par
In summary, foundation models are significantly transforming Sim2Real robotics research by embedding higher-level reasoning, semantic invariance, and generative diversity into learning pipelines. Early successes include language-driven visual invariance, diffusion-generated dynamic environments, and automated training schemes through LLMs, demonstrating substantial improvements in zero-shot transfer and sample efficiency. However, robustness, interpretability, and integration costs remain ongoing challenges. Future efforts are thus aimed both at extending foundation model capabilities and establishing rigorous safeguards through effective grounding, model compression, and validation to reliably utilize these models' potential for autonomous systems.

\section{Enabling Adaptable Robotic Systems}\label{sec:adaptable}
\begin{figure*}[!t]
    \centering
    \includegraphics[width=1\linewidth]{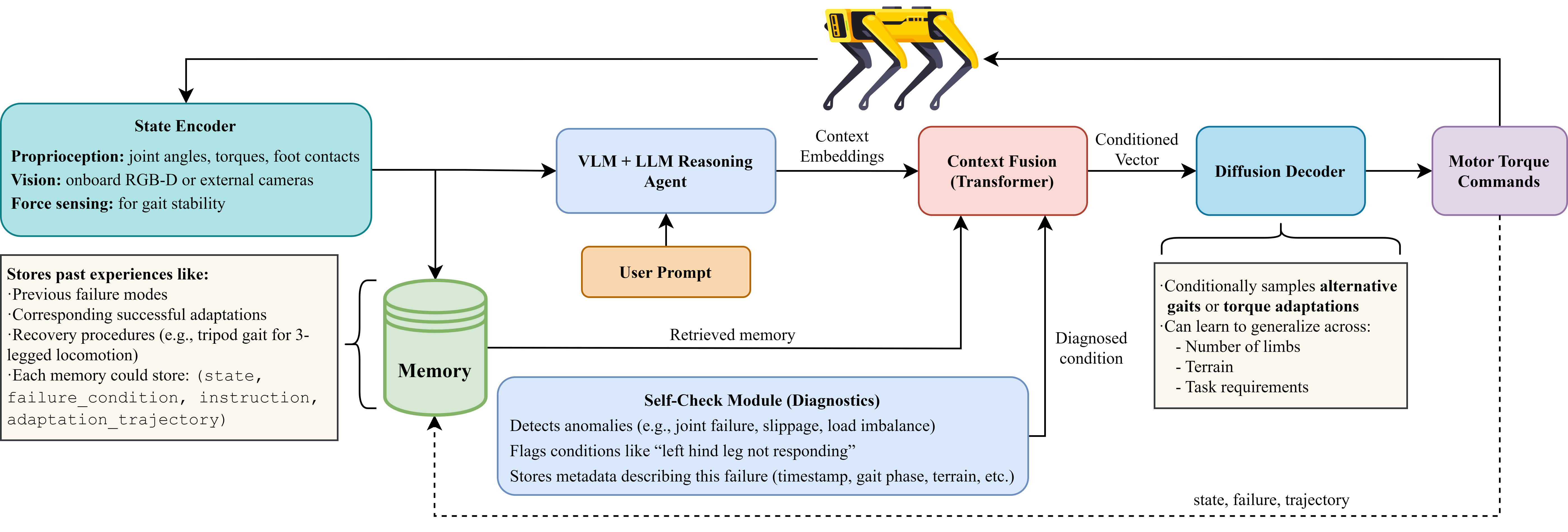}
    \caption{A generalized architecture for adaptive quadruped control that combines proprioceptive sensing, self-diagnostics, retrieval-augmented memory, and diffusion-based action generation. The LLM and VLM enable context-aware reasoning and support resilient behavior under failure conditions.}
    \label{fig:6}
\end{figure*}

An emerging direction involves leveraging foundation models to enhance robotic adaptability, enabling systems to handle unexpected changes in the environment or within themselves, such as damage or platform changes, without requiring extensive reprogramming. The high-level reasoning and code generation abilities of LLMs open the door to robots that can self-diagnose and adjust their strategies on the fly. Initial evidence for this can be seen in LLM-guided control frameworks, as discussed below. An example of such an architecture is shown in Fig.~\ref{fig:6}.
\subsection{LLM-Guided Controller Adaptation and Self-Repair}
Early work indicates that LLMs can serve as ``high-level brains'' that adapt a robot's low-level control on the fly. For example, Zahedifar et al. \cite{zahedifar2025llm} introduced a hybrid Lyapunov-LLM controller that achieved 100\% success in adapting a two-link manipulator to sudden dynamic changes, using CoT reasoning for real-time parameter tuning. Similar LLM-guided approaches have been applied to problems ranging from discrete controller synthesis and situated task planning to multibody dynamics modeling and humanoid locomotion control \cite{ishimizu2024towards, singh2023progprompt, gerstmayr2024multibody, sun2024leveraging}.
\par
However, a notable challenge is robustness. LLMs can occasionally propose unstable actions and incorrect fixes if their prompt context is insufficient. Ensuring physical safety and stability remains an active research area. Moreover, the compute overhead of querying an LLM can be nontrivial for real-time control. Recent work on lightweight specialized LLMs and optimization hints (e.g., embedding physical laws in the prompt) aims to mitigate latency \cite{sun2024leveraging}.

\subsection{Adaptive Planning and Strategy Regeneration}
Beyond low-level control, foundation models have shown effectiveness in adapting task plans on the fly when a robot’s initial strategy fails or new constraints arise. Skreta et al. \cite{skreta2024replan} introduced RePLan, a framework that combines vision and language models to enable online replanning in long-horizon tasks. Researchers have also integrated such adaptive planning into physical robot demonstrations on various platforms. Ouyang et al. \cite{ouyang2024long} utilized a quadrupedal robot system where multiple LLM-driven agents handle semantic planning, parameter calculation, and code synthesis, in conjunction with low-level RL skills, achieving an overall success rate of over 70\% in simulation.
\par
While adaptive planning with foundation models is still in its infancy, early critiques have emerged regarding reliability and computational efficiency. Current LLM planners operate at the level of high-level actions (taking seconds per decision), which is acceptable for task planning. However, making such adaptive reasoning fast enough for split-second reactions will require further optimization of these foundation models. Despite these caveats, the trend is clearly toward more integrated perception-reasoning-action loops. Drawing on massive prior knowledge, an LLM-based planner can often find a creative solution where a conventional planner gives up, and researchers are rapidly addressing the remaining failure modes through better grounding, safety constraints, and human feedback in training.

\subsection{Modular Architectures and Self-Assessment for Resilience}
A notable methodological trend is the move toward modular and multimodal foundation model architectures in robotics. Instead of relying on one monolithic model to do everything, researchers decompose the problem: separate pretrained models handle language reasoning, visual perception, and low-level control, with defined interfaces between them. This modular design not only mirrors the sense-plan-act pipeline of classical robotics but also improves transparency and adaptability.
The HiP framework \cite{ajay2023compositional}, developed at the Massachusetts Institute of Technology, is a prime example that uses three different foundation models arranged hierarchically. It uses a GPT-based language reasoner for high-level task planning, a video-based physics model to understand the environment, and an action model for motor control, each contributing its expertise to the overall plan. The planning process becomes an iterative loop in which the language model proposes a plan, the vision model imagines what that plan would look like in the physical world and refines it, and the action model calculates the motor sequences needed to execute it. The result is not only adaptability but also interpretability, as observers can inspect intermediate plans or visual predictions to understand what the robot ``thinks'' will happen. This makes it easier to diagnose failures when they occur. However, modular systems are only as strong as their weakest link. The HiP team pointed out that current vision foundation models (particularly for video-based planning) are not yet as powerful as their language counterparts, somewhat limiting the overall performance of the system.
\par
Similarly, memory enhancement is being explored to help robots learn over time. Instead of treating each task as isolated, the robot can maintain a history that an LLM can query when deciding a new course of action. Glocker et al. \cite{glocker2025llm} implemented this using Retrieval-Augmented Generation (RAG). The LLM-based planner retrieves timestamped records of past dialogue and actions, such as which objects were moved and where they were placed, and incorporates that context into its current reasoning. However, maintaining and querying memories raises both efficiency and reliability concerns. Too much information can overwhelm the context window of an LLM, while outdated or misleading memories may compromise the robot's performance. Research in this area is exploring how to retain salient memories and when to reset or update them.
\par
Across the four directions reviewed in Sections~\ref{sec:simulation}--\ref{sec:adaptable}, namely (i) simulation-driven robotics, (ii) robots in unstructured, open-world environments, (iii) bridging the Sim2Real gap, and (iv) enabling adaptable robotic systems, the common theme is an ambition to transcend the fragmented, brittle nature of current robot learning. By combining massive prior knowledge with tailored integration into simulations, real-world deployments, transfer learning pipelines, and adaptive control loops, respectively, foundation models have the potential to address the very gaps in current systems identified in Section~\ref{sec:holistic}. In short, robotics research is moving toward more holistic solutions where perception, reasoning, and action are unified by general models. It also aims to develop more resilient systems that handle novelty, uncertainty, and change gracefully, along with more scalable methodologies that leverage data and experience far beyond what individual labs can manually curate.

\section{Limitations and the Proposed Future Directions}\label{sec:future}
Despite the progress reviewed in Sections~\ref{sec:simulation}--\ref{sec:adaptable}, several fundamental limitations continue to constrain the reliable deployment of foundation models in real-world robotic systems. This section consolidates these limitations into six key areas and, for each area, outlines the future research directions proposed in this review to address them.

\subsection{Real-Time and Computational Constraints}
Current foundation models are computationally heavy and often too slow for onboard robotic deployment \cite{jeong2024survey}. Their high inference latency and memory footprint conflict with the strict real-time requirements of autonomous robots. This limitation impedes closed-loop control and rapid decision-making. Progress in model compression, distillation, and on-device optimization is needed to meet the latency and power constraints of robotic platforms.

\subsection{Data Scarcity and Physical Grounding}
Foundation models are primarily trained on internet-scale text or image data, with limited exposure to robot-specific information \cite{hu2023toward}. They fail to capture the long tail of physical scenarios and corner cases that are critical for reliable perception and control. This embodiment gap means that models may not yet fully understand real-world dynamics or sensor modalities. Bridging this gap will require extensive robot-relevant training data, obtained through simulation-generated experience and collaborative data collection, to ground models in physical reality.

\subsection{Limited Multimodal Integration}
Current foundation models excel primarily in language and vision but struggle to incorporate other modalities (e.g., tactile, audio, proprioceptive) that robots rely on \cite{jones2025beyond}. Truly holistic perception remains out of reach, as certain sensory data lack large, aligned datasets for model training. For example, 3D point clouds are vital in robotics but are challenging to use due to limited text-aligned data. Designing models that integrate diverse modalities without losing information is an open research frontier, which points to the need for new multimodal representations and cross-modal training paradigms.

\subsection{Safety and Reliability Concerns}
Integrating LLMs into robotics introduces significant safety risks. Models can hallucinate erroneous outputs or be manipulated by adversarial prompts, potentially leading to physically harmful actions \cite{xu2024hallucination, wu2024highlighting, ravichandran2025safety}. Traditional robot safety measures do not cover these failure modes, and current LLM safety techniques overlook the physical context of robotic operation. Ensuring reliable robot behavior will require new guardrails and validation layers (e.g., runtime safety filters, uncertainty monitors) to catch infeasible or dangerous commands before execution, along with robust training against distribution shifts and attacks.

\subsection{Lack of Interpretability and Transparency}
The decision-making process of foundation models is largely a black box, which poses challenges for debugging and trust. In safety-critical robotic tasks, the inability to explain why a model chose a certain action undermines user confidence and complicates validation. Addressing this issue requires research into explainable AI techniques and potentially hybrid architectures that make the robot’s internal reasoning more transparent. Early efforts (e.g., chain-of-thought prompting or modular policies) hint at more interpretable strategies \cite{zawalski2024robotic, ajay2023compositional}. However, developing robot policies that are truly understandable to humans without compromising performance remains an open challenge.

\subsection{Generalization and Robustness Limits}
Despite their broad training, today’s foundation models are not yet fully general-purpose in unstructured environments. They often fail to adapt to novel tasks, unseen environments, or changes in their own embodiment. For example, transferring knowledge across different robot morphologies or under extreme environmental conditions remains highly challenging. Models often exhibit weakness under distribution shifts, leading to performance drops outside the training distribution. Enhancing robust generalization will require advances in continual learning, domain adaptation, and possibly robotics-specific foundation models that learn from diverse embodied experiences. In short, foundation models have only partially alleviated the generalization problem, and truly resilient open-world robotic intelligence has yet to be achieved.

\section{Conclusions}\label{sec:conclusion}
This review has presented a structured synthesis of how foundation models, particularly LLMs and VLMs, are transforming robotics. It first traced the evolution of these models and examined their integration into the core robotic subsystems of perception, planning, control, and HRI, where they enable open-vocabulary perception, zero-shot task decomposition, code-based policy synthesis, and natural language interaction. It then analyzed four directions in which foundation models are being operationalized at the system level. In simulation-driven robotics, foundation models automate the generation of environments, tasks, and simulation code, and they support the coordination of heterogeneous multi-robot teams. In unstructured, open-world environments, they enable hierarchical task decomposition, generalist cross-embodiment policies, and predictive world models. To bridge the Sim2Real gap, they provide semantically aligned representations, generative domain adaptation, and automated reward and curriculum design. Finally, in adaptable robotic systems, they facilitate controller adaptation, online replanning, and memory-augmented modular architectures that enhance resilience.
\par
The analysis further indicates that, in their current form, foundation models are most effective when their semantic reasoning is coupled with physically grounded mechanisms that constrain their outputs to executable and safe actions, such as affordance-based feasibility checks, low-level controllers, and safety validation layers. Nevertheless, as discussed in Section~\ref{sec:future}, several limitations still hinder their reliable deployment, including high inference latency and computational cost, the scarcity of robot-specific data and the resulting lack of physical grounding, limited integration of modalities beyond language and vision, safety and reliability risks arising from hallucinations and adversarial prompts, limited interpretability, and insufficient robustness under distribution shifts and embodiment changes. The corresponding future research directions proposed in this review are model compression, distillation, and on-device optimization; large-scale robot-relevant data obtained through simulation and collaborative collection; new multimodal representations and cross-modal training paradigms; runtime safety guardrails and validation layers; explainable AI techniques and hybrid architectures; and continual learning, domain adaptation, and robotics-specific foundation models. Collectively, these directions outline a roadmap toward robotic systems that are more holistic, resilient, and scalable.
\par
This review has two main limitations. First, given the rapid pace of the field, models released after the literature survey are not covered, and some findings are drawn from preprints that may be revised. Second, the synthesis focuses primarily on LLMs and VLMs and is qualitative, since the heterogeneity of tasks, embodiments, and evaluation metrics across studies precludes direct quantitative comparison. Future work will therefore pursue a systematic, benchmark-based comparison of foundation model-driven robotic systems and extend the analysis to foundation models that incorporate additional sensing modalities.

\section*{Acknowledgment}
The authors used Claude (Anthropic) to improve the readability of this manuscript, reviewed all content carefully, and take full responsibility for the final version.

\bibliographystyle{IEEEtran}
\bibliography{References}

@article{kim2024survey,
  title={A survey on integration of large language models with intelligent robots},
  author={Kim, Yeseung and Kim, Dohyun and Choi, Jieun and Park, Jisang and Oh, Nayoung and Park, Daehyung},
  journal={Intelligent Service Robotics},
  volume={17},
  number={5},
  pages={1091--1107},
  year={2024},
  publisher={Springer}
}

@article{brown2020language,
  title={Language models are few-shot learners},
  author={Brown, Tom and Mann, Benjamin and Ryder, Nick and Subbiah, Melanie and Kaplan, Jared D and Dhariwal, Prafulla and Neelakantan, Arvind and Shyam, Pranav and Sastry, Girish and Askell, Amanda and others},
  journal={Advances in Neural Information Processing Systems},
  volume={33},
  pages={1877--1901},
  year={2020}
}

@misc{narang2022pathways,
  title={Pathways language model ({PaLM}): Scaling to 540 billion parameters for breakthrough performance},
  author={Narang, Sharan and Chowdhery, Aakanksha},
  howpublished={Google Research Blog},
  year={2022},
  url={https://research.google/blog/pathways-language-model-palm-scaling-to-540-billion-parameters-for-breakthrough-performance/}
}

@article{sanderson2023gpt,
  title={{GPT-4} is here: What scientists think},
  author={Sanderson, Katharine},
  journal={Nature},
  volume={615},
  number={7954},
  pages={773},
  year={2023},
  publisher={Nature}
}

@article{dorigo2021swarm,
  title={Swarm robotics: Past, present, and future [point of view]},
  author={Dorigo, Marco and Theraulaz, Guy and Trianni, Vito},
  journal={Proceedings of the IEEE},
  volume={109},
  number={7},
  pages={1152--1165},
  year={2021},
  publisher={IEEE}
}

@article{karoly2020deep,
  title={Deep learning in robotics: Survey on model structures and training strategies},
  author={K{\'a}roly, Art{\'u}r Istv{\'a}n and Galambos, P{\'e}ter and Kuti, J{\'o}zsef and Rudas, Imre J},
  journal={IEEE Transactions on Systems, Man, and Cybernetics: Systems},
  volume={51},
  number={1},
  pages={266--279},
  year={2021},
  publisher={IEEE}
}

@article{coombs2020strategic,
  title={The strategic impacts of Intelligent Automation for knowledge and service work: An interdisciplinary review},
  author={Coombs, Crispin and Hislop, Donald and Taneva, Stanimira K and Barnard, Sarah},
  journal={The Journal of Strategic Information Systems},
  volume={29},
  number={4},
  pages={101600},
  year={2020},
  publisher={Elsevier}
}

@inproceedings{ahn2022can,
  title={Do as {I} can, not as {I} say: Grounding language in robotic affordances},
  author={Ichter, Brian and Brohan, Anthony and Chebotar, Yevgen and Finn, Chelsea and Hausman, Karol and Herzog, Alexander and Ho, Daniel and Ibarz, Julian and Irpan, Alex and Jang, Eric and others},
  booktitle={Proceedings of the 6th Conference on Robot Learning (CoRL)},
  series={Proceedings of Machine Learning Research},
  volume={205},
  pages={287--318},
  year={2023},
  publisher={PMLR}
}

@article{guan2023leveraging,
  title={Leveraging pre-trained large language models to construct and utilize world models for model-based task planning},
  author={Guan, Lin and Valmeekam, Karthik and Sreedharan, Sarath and Kambhampati, Subbarao},
  journal={Advances in Neural Information Processing Systems},
  volume={36},
  pages={79081--79094},
  year={2023}
}

@inproceedings{liang2023code,
  title={Code as policies: Language model programs for embodied control},
  author={Liang, Jacky and Huang, Wenlong and Xia, Fei and Xu, Peng and Hausman, Karol and Ichter, Brian and Florence, Pete and Zeng, Andy},
  booktitle={2023 IEEE International Conference on Robotics and Automation (ICRA)},
  pages={9493--9500},
  year={2023},
  organization={IEEE}
}

@article{wang2024large,
  title={Large language models for robotics: Opportunities, challenges, and perspectives},
  author={Wang, Jiaqi and Shi, Enze and Hu, Huawen and Ma, Chong and Liu, Yiheng and Wang, Xuhui and Yao, Yincheng and Liu, Xuan and Ge, Bao and Zhang, Shu},
  journal={Journal of Automation and Intelligence},
  volume={4},
  number={1},
  pages={52--64},
  year={2025},
  doi={10.1016/j.jai.2024.12.003}
}

@article{mu2023embodiedgpt,
  title={{EmbodiedGPT}: Vision-language pre-training via embodied chain of thought},
  author={Mu, Yao and Zhang, Qinglong and Hu, Mengkang and Wang, Wenhai and Ding, Mingyu and Jin, Jun and Wang, Bin and Dai, Jifeng and Qiao, Yu and Luo, Ping},
  journal={Advances in Neural Information Processing Systems},
  volume={36},
  pages={25081--25094},
  year={2023}
}

@article{liu2024enhancing,
  title={Enhancing robotic manipulation with {AI} feedback from multimodal large language models},
  author={Liu, Jinyi and Yuan, Yifu and Hao, Jianye and Ni, Fei and Fu, Lingzhi and Chen, Yibin and Zheng, Yan},
  journal={arXiv preprint arXiv:2402.14245},
  year={2024}
}

@article{ding2023integrating,
  title={Integrating action knowledge and {LLMs} for task planning and situation handling in open worlds},
  author={Ding, Yan and Zhang, Xiaohan and Amiri, Saeid and Cao, Nieqing and Yang, Hao and Kaminski, Andy and Esselink, Chad and Zhang, Shiqi},
  journal={Autonomous Robots},
  volume={47},
  number={8},
  pages={981--997},
  year={2023},
  publisher={Springer}
}

@article{zeng2023large,
  title={Large language models for robotics: A survey},
  author={Zeng, Fanlong and Gan, Wensheng and Huai, Zezheng and Sun, Lichao and Chen, Hechang and Wang, Yongheng and Liu, Ning and Yu, Philip S},
  journal={arXiv preprint arXiv:2311.07226},
  year={2023}
}

@article{li2025large,
  title={Large language models for multi-robot systems: A survey},
  author={Li, Peihan and An, Zijian and Abrar, Shams and Zhou, Lifeng},
  journal={Autonomous Robots},
  volume={50},
  number={3},
  pages={30},
  year={2026},
  doi={10.1007/s10514-026-10257-4}
}

@article{floridi2020gpt,
  title={{GPT-3}: Its nature, scope, limits, and consequences},
  author={Floridi, Luciano and Chiriatti, Massimo},
  journal={Minds and Machines},
  volume={30},
  number={4},
  pages={681--694},
  year={2020},
  publisher={Springer}
}

@inproceedings{devlin2019bert,
  title={{BERT}: Pre-training of deep bidirectional transformers for language understanding},
  author={Devlin, Jacob and Chang, Ming-Wei and Lee, Kenton and Toutanova, Kristina},
  booktitle={Proceedings of the 2019 Conference of the North American Chapter of the Association for Computational Linguistics: Human Language Technologies, Volume 1 (Long and Short Papers)},
  pages={4171--4186},
  year={2019}
}

@article{raffel2020exploring,
  title={Exploring the limits of transfer learning with a unified text-to-text transformer},
  author={Raffel, Colin and Shazeer, Noam and Roberts, Adam and Lee, Katherine and Narang, Sharan and Matena, Michael and Zhou, Yanqi and Li, Wei and Liu, Peter J},
  journal={Journal of Machine Learning Research},
  volume={21},
  number={140},
  pages={1--67},
  year={2020}
}

@article{chowdhery2023palm,
  title={{PaLM}: Scaling language modeling with pathways},
  author={Chowdhery, Aakanksha and Narang, Sharan and Devlin, Jacob and Bosma, Maarten and Mishra, Gaurav and Roberts, Adam and Barham, Paul and Chung, Hyung Won and Sutton, Charles and Gehrmann, Sebastian and others},
  journal={Journal of Machine Learning Research},
  volume={24},
  number={240},
  pages={1--113},
  year={2023}
}

@article{achiam2023gpt,
  title={{GPT-4} technical report},
  author={Achiam, Josh and Adler, Steven and Agarwal, Sandhini and Ahmad, Lama and Akkaya, Ilge and Aleman, Florencia Leoni and Almeida, Diogo and Altenschmidt, Janko and Altman, Sam and Anadkat, Shyamal and others},
  journal={arXiv preprint arXiv:2303.08774},
  year={2023}
}

@article{touvron2023llama,
  title={{LLaMA}: Open and efficient foundation language models},
  author={Touvron, Hugo and Lavril, Thibaut and Izacard, Gautier and Martinet, Xavier and Lachaux, Marie-Anne and Lacroix, Timoth{\'e}e and Rozi{\`e}re, Baptiste and Goyal, Naman and Hambro, Eric and Azhar, Faisal and others},
  journal={arXiv preprint arXiv:2302.13971},
  year={2023}
}

@article{yang2023dawn,
  title={The dawn of {LMMs}: Preliminary explorations with {GPT-4V(ision)}},
  author={Yang, Zhengyuan and Li, Linjie and Lin, Kevin and Wang, Jianfeng and Lin, Chung-Ching and Liu, Zicheng and Wang, Lijuan},
  journal={arXiv preprint arXiv:2309.17421},
  year={2023}
}

@article{lu2019vilbert,
  title={{ViLBERT}: Pretraining task-agnostic visiolinguistic representations for vision-and-language tasks},
  author={Lu, Jiasen and Batra, Dhruv and Parikh, Devi and Lee, Stefan},
  journal={Advances in Neural Information Processing Systems},
  volume={32},
  year={2019}
}

@inproceedings{radford2021learning,
  title={Learning transferable visual models from natural language supervision},
  author={Radford, Alec and Kim, Jong Wook and Hallacy, Chris and Ramesh, Aditya and Goh, Gabriel and Agarwal, Sandhini and Sastry, Girish and Askell, Amanda and Mishkin, Pamela and Clark, Jack and others},
  booktitle={International Conference on Machine Learning},
  pages={8748--8763},
  year={2021},
  organization={PMLR},
}

@inproceedings{wang2021simvlm,
  title={{SimVLM}: Simple visual language model pretraining with weak supervision},
  author={Wang, Zirui and Yu, Jiahui and Yu, Adams Wei and Dai, Zihang and Tsvetkov, Yulia and Cao, Yuan},
  booktitle={Proceedings of the International Conference on Learning Representations (ICLR)},
  year={2022}
}

@article{alayrac2022flamingo,
  title={Flamingo: A visual language model for few-shot learning},
  author={Alayrac, Jean-Baptiste and Donahue, Jeff and Luc, Pauline and Miech, Antoine and Barr, Iain and Hasson, Yana and Lenc, Karel and Mensch, Arthur and Millican, Katherine and Reynolds, Malcolm and others},
  journal={Advances in Neural Information Processing Systems},
  volume={35},
  pages={23716--23736},
  year={2022}
}

@inproceedings{li2023blip,
  title={{BLIP-2}: Bootstrapping language-image pre-training with frozen image encoders and large language models},
  author={Li, Junnan and Li, Dongxu and Savarese, Silvio and Hoi, Steven},
  booktitle={International Conference on Machine Learning},
  pages={19730--19742},
  year={2023},
  organization={PMLR}
}

@article{chen2021evaluating,
  title={Evaluating large language models trained on code},
  author={Chen, Mark and Tworek, Jerry and Jun, Heewoo and Yuan, Qiming and Pinto, Henrique Ponde De Oliveira and Kaplan, Jared and Edwards, Harri and Burda, Yuri and Joseph, Nicholas and Brockman, Greg and others},
  journal={arXiv preprint arXiv:2107.03374},
  year={2021}
}

@article{reed2022generalist,
  title={A generalist agent},
  author={Reed, Scott and Zolna, Konrad and Parisotto, Emilio and Colmenarejo, Sergio G{\'o}mez and Novikov, Alexander and Barth-Maron, Gabriel and Gim{\'e}nez, Mai and Sulsky, Yury and Kay, Jackie and Springenberg, Jost Tobias and others},
  journal={Transactions on Machine Learning Research},
  year={2022}
}

@article{ouyang2022training,
  title={Training language models to follow instructions with human feedback},
  author={Ouyang, Long and Wu, Jeffrey and Jiang, Xu and Almeida, Diogo and Wainwright, Carroll and Mishkin, Pamela and Zhang, Chong and Agarwal, Sandhini and Slama, Katarina and Ray, Alex and others},
  journal={Advances in Neural Information Processing Systems},
  volume={35},
  pages={27730--27744},
  year={2022}
}

@article{yoshida2023textmotiongroundinggpt4,
  title={From text to motion: Grounding {GPT-4} in a humanoid robot `{Alter3}'},
  author={Yoshida, Takahide and Masumori, Atsushi and Ikegami, Takashi},
  journal={Frontiers in Robotics and AI},
  volume={12},
  pages={1581110},
  year={2025},
  doi={10.3389/frobt.2025.1581110}
}

@article{o2025exploring,
  title={Exploring {GPT-4} for robotic agent strategy with real-time state feedback and a reactive behaviour framework},
  author={O'Brien, Thomas and Sims, Ysobel},
  journal={arXiv preprint arXiv:2503.23601},
  year={2025}
}

@inproceedings{stark2024dobby,
  title={Dobby: A conversational service robot driven by {GPT-4}},
  author={Stark, Carson and Chun, Bohkyung and Charleston, Casey and Ravi, Varsha and Pabon, Luis and Sunkari, Surya and Mohan, Tarun and Stone, Peter and Hart, Justin},
  booktitle={2024 33rd IEEE International Conference on Robot and Human Interactive Communication (ROMAN)},
  pages={1362--1369},
  year={2024},
  organization={IEEE}
}

@article{barkley2025semantic,
  title={Semantic intelligence: Integrating {GPT-4} with {A}* planning in low-cost robotics},
  author={Barkley, Jesse and George, Abraham and Farimani, Amir Barati},
  journal={arXiv preprint arXiv:2505.01931},
  year={2025}
}

@article{koubaa2025next,
  title={Next-generation human-robot interaction with {ChatGPT} and {Robot Operating System}},
  author={Koubaa, Anis and Ammar, Adel and Boulila, Wadii},
  journal={Software: Practice and Experience},
  volume={55},
  number={2},
  pages={355--382},
  year={2025},
  publisher={Wiley Online Library}
}

@article{kawaharazuka2024robotic,
  title={Robotic environmental state recognition with pre-trained vision-language models and black-box optimization},
  author={Kawaharazuka, Kento and Obinata, Yoshiki and Kanazawa, Naoaki and Okada, Kei and Inaba, Masayuki},
  journal={Advanced Robotics},
  volume={38},
  number={18},
  pages={1255--1264},
  year={2024},
  publisher={Taylor \& Francis}
}

@inproceedings{shafiullah2022clip,
  title={{CLIP-Fields}: Weakly supervised semantic fields for robotic memory},
  author={Shafiullah, Nur Muhammad Mahi and Paxton, Chris and Pinto, Lerrel and Chintala, Soumith and Szlam, Arthur},
  booktitle={Proceedings of Robotics: Science and Systems (RSS)},
  year={2023},
  doi={10.15607/RSS.2023.XIX.074}
}

@inproceedings{nguyen2024robotic,
  title={{Robotic-CLIP}: Fine-tuning {CLIP} on action data for robotic applications},
  author={Nguyen, Nghia and Vu, Minh Nhat and Ta, Tung D. and Huang, Baoru and Vo, Thieu and Le, Ngan and Nguyen, Anh},
  booktitle={Proceedings of the IEEE International Conference on Robotics and Automation (ICRA)},
  pages={5930--5936},
  year={2025},
  doi={10.1109/ICRA55743.2025.11127829}
}

@inproceedings{kang2024clip,
  title={{CLIP-RT}: Learning language-conditioned robotic policies from natural language supervision},
  author={Kang, Gi-Cheon and Kim, Junghyun and Shim, Kyuhwan and Lee, Jun Ki and Zhang, Byoung-Tak},
  booktitle={Proceedings of Robotics: Science and Systems (RSS)},
  year={2025},
  doi={10.15607/RSS.2025.XXI.016}
}

@article{shibata2024clip,
  title={{CLIP} feature-based randomized control using images and text for multiple tasks and robots},
  author={Shibata, Kazuki and Deguchi, Hideki and Taguchi, Shun},
  journal={Advanced Robotics},
  volume={38},
  number={15},
  pages={1066--1078},
  year={2024},
  publisher={Taylor \& Francis}
}

@article{sontakke2023roboclip,
  title={{RoboCLIP}: One demonstration is enough to learn robot policies},
  author={Sontakke, Sumedh and Zhang, Jesse and Arnold, S{\'e}b and Pertsch, Karl and B{\i}y{\i}k, Erdem and Sadigh, Dorsa and Finn, Chelsea and Itti, Laurent},
  journal={Advances in Neural Information Processing Systems},
  volume={36},
  pages={55681--55693},
  year={2023}
}

@article{luo2024transformer,
  title={Transformer-based vision-language alignment for robot navigation and question answering},
  author={Luo, Haonan and Guo, Ziyu and Wu, Zhenyu and Teng, Fei and Li, Tianrui},
  journal={Information Fusion},
  volume={108},
  pages={102351},
  year={2024},
  publisher={Elsevier}
}

@article{tavassoli2023expanding,
  title={Expanding Frozen Vision-Language Models without Retraining: Towards Improved Robot Perception},
  author={Tavassoli, Riley and Amani, Mani and Akhavian, Reza},
  journal={arXiv preprint arXiv:2308.16493},
  year={2023}
}

@article{wang2025roboflamingo,
  title={{RoboFlamingo-Plus}: Fusion of depth and {RGB} perception with vision-language models for enhanced robotic manipulation},
  author={Wang, Sheng},
  journal={arXiv preprint arXiv:2503.19510},
  year={2025}
}

@inproceedings{du2023vision,
  title={Vision-language models as success detectors},
  author={Du, Yuqing and Konyushkova, Ksenia and Denil, Misha and Raju, Akhil and Landon, Jessica and Hill, Felix and de Freitas, Nando and Cabi, Serkan},
  booktitle={Proceedings of the 2nd Conference on Lifelong Learning Agents (CoLLAs)},
  series={Proceedings of Machine Learning Research},
  volume={232},
  pages={120--136},
  year={2023},
  publisher={PMLR}
}

@inproceedings{zhou2024llm,
  title={{LLM-BT}: Performing robotic adaptive tasks based on large language models and behavior trees},
  author={Zhou, Haotian and Lin, Yunhan and Yan, Longwu and Zhu, Jihong and Min, Huasong},
  booktitle={2024 IEEE International Conference on Robotics and Automation (ICRA)},
  pages={16655--16661},
  year={2024},
  organization={IEEE}
}

@inproceedings{silva2024llm,
  title={{LLM} text generation for service robot context},
  author={Silva, L{\'a}zaro Q and Mascarenhas, Ana Patricia FM and Sim{\~o}es, Marco AC and Rodowanski, Ivano{\'e} J and De Campos, Jorge Alberto P and De Souza, Josemar Rodrigues and Da Silva Filho, Jose Grimaldo},
  booktitle={2024 Brazilian Symposium on Robotics (SBR), and 2024 Workshop on Robotics in Education (WRE)},
  pages={25--30},
  year={2024},
  organization={IEEE}
}

@article{latif20243p,
  title={{3P-LLM}: Probabilistic path planning using large language model for autonomous robot navigation},
  author={Latif, Ehsan},
  journal={arXiv preprint arXiv:2403.18778},
  year={2024}
}

@article{zhao2024applying,
  title={Applying Large Language Model to a Control System for Multi-Robot Task Assignment},
  author={Zhao, Wen and Li, Liqiao and Zhan, Hanwen and Wang, Yingqi and Fu, Yiqi},
  journal={Drones},
  volume={8},
  number={12},
  pages={728},
  year={2024},
  publisher={MDPI}
}

@article{gao2024integrating,
  title={Integrating Large Language Model for Natural Language-Based Instruction toward Robust Human-Robot Collaboration},
  author={Gao, Fanru and Xia, Liqiao and Zhang, Jianjing and Liu, Sichao and Wang, Lihui and Gao, Robert X},
  journal={Procedia CIRP},
  volume={130},
  pages={313--318},
  year={2024},
  publisher={Elsevier}
}

@inproceedings{zhang2023large,
  title={Large language models as zero-shot human models for human-robot interaction},
  author={Zhang, Bowen and Soh, Harold},
  booktitle={2023 IEEE/RSJ International Conference on Intelligent Robots and Systems (IROS)},
  pages={7961--7968},
  year={2023},
  organization={IEEE}
}

@article{firoozi2025foundation,
  title={Foundation models in robotics: Applications, challenges, and the future},
  author={Firoozi, Roya and Tucker, Johnathan and Tian, Stephen and Majumdar, Anirudha and Sun, Jiankai and Liu, Weiyu and Zhu, Yuke and Song, Shuran and Kapoor, Ashish and Hausman, Karol and others},
  journal={The International Journal of Robotics Research},
  volume={44},
  number={5},
  pages={701--739},
  year={2025},
  publisher={SAGE Publications Sage UK: London, England}
}

@article{shu2024llms,
  title={{LLMs} for coding and robotics education},
  author={Shu, Peng and Zhao, Huaqin and Jiang, Hanqi and Li, Yiwei and Xu, Shaochen and Pan, Yi and Wu, Zihao and Liu, Zhengliang and Lu, Guoyu and Guan, Le and others},
  journal={arXiv preprint arXiv:2402.06116},
  year={2024}
}

@inproceedings{mu2024robocodex,
  title={{RoboCodeX}: Multimodal code generation for robotic behavior synthesis},
  author={Mu, Yao and Chen, Junting and Zhang, Qing-Long and Chen, Shoufa and Yu, Qiaojun and Ge, Chongjian and Chen, Runjian and Liang, Zhixuan and Hu, Mengkang and Tao, Chaofan and others},
  booktitle={Proceedings of the 41st International Conference on Machine Learning (ICML)},
  series={Proceedings of Machine Learning Research},
  volume={235},
  pages={36434--36454},
  year={2024},
  publisher={PMLR}
}

@inproceedings{sikorski2025deployment,
  title={Deployment of Large Language Models to Control Mobile Robots at the Edge},
  author={Sikorski, Pascal and Schrader, Leendert and Yu, Kaleb and Billadeau, Lucy and Meenakshi, Jinka and Mutharasan, Naveena and Esposito, Flavio and AliAkbarpour, Hadi and Babaiasl, Madi},
  booktitle={2025 3rd International Conference on Mechatronics, Control and Robotics (ICMCR)},
  pages={19--24},
  year={2025},
  organization={IEEE}
}

@inproceedings{macdonald2024language,
  title={Language, camera, autonomy! {P}rompt-engineered robot control for rapidly evolving deployment},
  author={Macdonald, Jacob P and Mallick, Rohit and Wollaber, Allan B and Pe{\~n}a, Jaime D and McNeese, Nathan and Siu, Ho Chit},
  booktitle={Companion of the 2024 ACM/IEEE International Conference on Human-Robot Interaction},
  pages={717--721},
  year={2024}
}

@inproceedings{li2024manipllm,
  title={{ManipLLM}: Embodied multimodal large language model for object-centric robotic manipulation},
  author={Li, Xiaoqi and Zhang, Mingxu and Geng, Yiran and Geng, Haoran and Long, Yuxing and Shen, Yan and Zhang, Renrui and Liu, Jiaming and Dong, Hao},
  booktitle={Proceedings of the IEEE/CVF Conference on Computer Vision and Pattern Recognition},
  pages={18061--18070},
  year={2024}
}

@article{liu2024robomamba,
  title={{RoboMamba}: Efficient vision-language-action model for robotic reasoning and manipulation},
  author={Liu, Jiaming and Liu, Mengzhen and Wang, Zhenyu and An, Pengju and Li, Xiaoqi and Zhou, Kaichen and Yang, Senqiao and Zhang, Renrui and Guo, Yandong and Zhang, Shanghang},
  journal={Advances in Neural Information Processing Systems},
  volume={37},
  pages={40085--40110},
  year={2024}
}

@inproceedings{kong2024embodied,
  title={Embodied {AI} in mobile robot simulation with {EyeSim}: Coverage path planning with large language models},
  author={Kong, Xiangrui and Zhang, Wenxiao and Hong, Jin and Br{\"a}unl, Thomas},
  booktitle={Proceedings of the International Conference on Simulation and Modeling Methodologies, Technologies and Applications (SIMULTECH)},
  pages={185--192},
  year={2025},
  doi={10.5220/0013455100003970}
}

@article{sobo2025evaluating,
  title={Evaluating {LLMs} for code generation in {HRI}: A comparative study of {ChatGPT}, {Gemini}, and {Claude}},
  author={Sobo, Andrei and Mubarak, Awes and Baimagambetov, Almas and Polatidis, Nikolaos},
  journal={Applied Artificial Intelligence},
  volume={39},
  number={1},
  pages={2439610},
  year={2025},
  publisher={Taylor \& Francis}
}

@inproceedings{liu2023improvedllava,
  title={Improved baselines with visual instruction tuning},
  author={Liu, Haotian and Li, Chunyuan and Li, Yuheng and Lee, Yong Jae},
  booktitle={Proceedings of the IEEE/CVF Conference on Computer Vision and Pattern Recognition (CVPR)},
  pages={26286--26296},
  year={2024},
  doi={10.1109/CVPR52733.2024.02484}
}

@inproceedings{jin2024reasoning,
  title={Reasoning grasping via multimodal large language model},
  author={Jin, Shiyu and Xu, Jinxuan and Lei, Yutian and Zhang, Liangjun},
  booktitle={Proceedings of the 8th Conference on Robot Learning (CoRL)},
  series={Proceedings of Machine Learning Research},
  volume={270},
  pages={3809--3827},
  year={2025},
  publisher={PMLR}
}

@inproceedings{chen2025robo2vlm,
  title={{Robo2VLM}: Improving visual question answering using large-scale robot manipulation data},
  author={Chen, Kaiyuan Eric and Xie, Shuangyu and Ma, Zehan and Sanketi, Pannag and Goldberg, Ken},
  booktitle={Advances in Neural Information Processing Systems (NeurIPS)},
  volume={38},
  pages={24644--24679},
  year={2025},
  doi={10.52202/085713-0733}
}

@article{caesar2024enabling,
  title={Enabling action crossmodality for a pretrained large language model},
  author={Caesar, Anton and {\"O}zdemir, Ozan and Weber, Cornelius and Wermter, Stefan},
  journal={Natural Language Processing Journal},
  volume={7},
  pages={100072},
  year={2024},
  publisher={Elsevier}
}

@inproceedings{kannan2024smart,
  title={{SMART-LLM}: Smart multi-agent robot task planning using large language models},
  author={Kannan, Shyam Sundar and Venkatesh, Vishnunandan LN and Min, Byung-Cheol},
  booktitle={2024 IEEE/RSJ International Conference on Intelligent Robots and Systems (IROS)},
  pages={12140--12147},
  year={2024},
  organization={IEEE}
}

@article{wei2022emergent,
  title={Emergent abilities of large language models},
  author={Wei, Jason and Tay, Yi and Bommasani, Rishi and Raffel, Colin and Zoph, Barret and Borgeaud, Sebastian and Yogatama, Dani and Bosma, Maarten and Zhou, Denny and Metzler, Donald and others},
  journal={Transactions on Machine Learning Research},
  year={2022}
}

@inproceedings{gu2023llm,
  title={{LLM}-based code generation method for {Golang} compiler testing},
  author={Gu, Qiuhan},
  booktitle={Proceedings of the 31st ACM Joint European Software Engineering Conference and Symposium on the Foundations of Software Engineering},
  pages={2201--2203},
  year={2023}
}

@inproceedings{huang2022language,
  title={Language models as zero-shot planners: Extracting actionable knowledge for embodied agents},
  author={Huang, Wenlong and Abbeel, Pieter and Pathak, Deepak and Mordatch, Igor},
  booktitle={International Conference on Machine Learning},
  pages={9118--9147},
  year={2022},
  organization={PMLR}
}

@inproceedings{huang2022inner,
  title={Inner monologue: Embodied reasoning through planning with language models},
  author={Huang, Wenlong and Xia, Fei and Xiao, Ted and Chan, Harris and Liang, Jacky and Florence, Pete and Zeng, Andy and Tompson, Jonathan and Mordatch, Igor and Chebotar, Yevgen and others},
  booktitle={Proceedings of the 6th Conference on Robot Learning (CoRL)},
  series={Proceedings of Machine Learning Research},
  volume={205},
  pages={1769--1782},
  year={2023},
  publisher={PMLR}
}

@inproceedings{song2023llm,
  title={{LLM-Planner}: Few-shot grounded planning for embodied agents with large language models},
  author={Song, Chan Hee and Wu, Jiaman and Washington, Clayton and Sadler, Brian M and Chao, Wei-Lun and Su, Yu},
  booktitle={Proceedings of the IEEE/CVF International Conference on Computer Vision},
  pages={2998--3009},
  year={2023}
}

@article{gestrin2024nl2plan,
  title={{NL2Plan}: Robust {LLM}-driven planning from minimal text descriptions},
  author={Gestrin, Elliot and Kuhlmann, Marco and Seipp, Jendrik},
  journal={arXiv preprint arXiv:2405.04215},
  year={2024}
}

@inproceedings{allgeuer2024robots,
  title={When robots get chatty: Grounding multimodal human-robot conversation and collaboration},
  author={Allgeuer, Philipp and Ali, Hassan and Wermter, Stefan},
  booktitle={International Conference on Artificial Neural Networks},
  pages={306--321},
  year={2024},
  organization={Springer}
}

@inproceedings{kim2024understanding,
  title={Understanding large-language model ({LLM})-powered human-robot interaction},
  author={Kim, Callie Y and Lee, Christine P and Mutlu, Bilge},
  booktitle={Proceedings of the 2024 ACM/IEEE International Conference on Human-Robot Interaction},
  pages={371--380},
  year={2024}
}

@inproceedings{lykov2024llm,
  title={{LLM-BRAIn}: {AI}-driven fast generation of robot behaviour tree based on large language model},
  author={Lykov, Artem and Tsetserukou, Dzmitry},
  booktitle={2024 2nd International Conference on Foundation and Large Language Models (FLLM)},
  pages={392--397},
  year={2024},
  organization={IEEE}
}

@article{li2024personal,
  title={Personal {LLM} agents: Insights and survey about the capability, efficiency and security},
  author={Li, Yuanchun and Wen, Hao and Wang, Weijun and Li, Xiangyu and Yuan, Yizhen and Liu, Guohong and Liu, Jiacheng and Xu, Wenxing and Wang, Xiang and Sun, Yi and others},
  journal={arXiv preprint arXiv:2401.05459},
  year={2024}
}

@inproceedings{luu2024questioning,
  title={Questioning, Answering, and Captioning for Zero-Shot Detailed Image Caption},
  author={Luu, Duc-Tuan and Le, Viet-Tuan and Vo, Duc Minh},
  booktitle={Proceedings of the Asian Conference on Computer Vision},
  pages={242--259},
  year={2024}
}

@article{liu2024right,
  title={Right this way: Can {VLMs} guide us to see more to answer questions?},
  author={Liu, Li and Yang, Diji and Zhong, Sijia and Tholeti, Kalyana Suma Sree and Ding, Lei and Zhang, Yi and Gilpin, Leilani},
  journal={Advances in Neural Information Processing Systems},
  volume={37},
  pages={132946--132976},
  year={2024}
}

@article{hong2024cogvlm2,
  title={{CogVLM2}: Visual language models for image and video understanding},
  author={Hong, Wenyi and Wang, Weihan and Ding, Ming and Yu, Wenmeng and Lv, Qingsong and Wang, Yan and Cheng, Yean and Huang, Shiyu and Ji, Junhui and Xue, Zhao and others},
  journal={arXiv preprint arXiv:2408.16500},
  year={2024}
}

@inproceedings{gao2024physically,
  title={Physically grounded vision-language models for robotic manipulation},
  author={Gao, Jensen and Sarkar, Bidipta and Xia, Fei and Xiao, Ted and Wu, Jiajun and Ichter, Brian and Majumdar, Anirudha and Sadigh, Dorsa},
  booktitle={2024 IEEE International Conference on Robotics and Automation (ICRA)},
  pages={12462--12469},
  year={2024},
  organization={IEEE}
}

@article{li2024lmeye,
  title={{LMEye}: An interactive perception network for large language models},
  author={Li, Yunxin and Hu, Baotian and Chen, Xinyu and Ma, Lin and Xu, Yong and Zhang, Min},
  journal={IEEE Transactions on Multimedia},
  volume={26},
  pages={10952--10964},
  year={2024},
  doi={10.1109/TMM.2024.3428317}
}

@inproceedings{zhao2023chat,
  title={Chat with the environment: Interactive multimodal perception using large language models},
  author={Zhao, Xufeng and Li, Mengdi and Weber, Cornelius and Hafez, Muhammad Burhan and Wermter, Stefan},
  booktitle={2023 IEEE/RSJ International Conference on Intelligent Robots and Systems (IROS)},
  pages={3590--3596},
  year={2023},
  organization={IEEE}
}

@article{vaswani2017attention,
  title={Attention is all you need},
  author={Vaswani, Ashish and Shazeer, Noam and Parmar, Niki and Uszkoreit, Jakob and Jones, Llion and Gomez, Aidan N and Kaiser, {\L}ukasz and Polosukhin, Illia},
  journal={Advances in Neural Information Processing Systems},
  volume={30},
  year={2017}
}

@book{xie2003fundamentals,
  title={Fundamentals of Robotics: Linking Perception to Action},
  author={Xie, Ming},
  series={Series in Machine Perception and Artificial Intelligence},
  volume={54},
  publisher={World Scientific},
  address={Singapore},
  year={2003},
  doi={10.1142/5230}
}

@article{seminara2019active,
  title={Active haptic perception in robots: A review},
  author={Seminara, Lucia and Gastaldo, Paolo and Watt, Simon J and Valyear, Kenneth F and Zuher, Fernando and Mastrogiovanni, Fulvio},
  journal={Frontiers in Neurorobotics},
  volume={13},
  pages={53},
  year={2019},
  publisher={Frontiers Media SA}
}

@article{galceran2013survey,
  title={A survey on coverage path planning for robotics},
  author={Galceran, Enric and Carreras, Marc},
  journal={Robotics and Autonomous Systems},
  volume={61},
  number={12},
  pages={1258--1276},
  year={2013},
  publisher={Elsevier}
}

@book{kurdila2019dynamics,
  title={Dynamics and Control of Robotic Systems},
  author={Kurdila, Andrew J. and Ben-Tzvi, Pinhas},
  publisher={Wiley},
  address={Hoboken, NJ, USA},
  year={2019}
}

@article{sheridan2016human,
  title={Human--robot interaction: Status and challenges},
  author={Sheridan, Thomas B},
  journal={Human Factors},
  volume={58},
  number={4},
  pages={525--532},
  year={2016},
  publisher={SAGE Publications Sage CA: Los Angeles, CA}
}

@article{ren2024deep,
  title={Deep learning and machine learning---{O}bject detection and semantic segmentation: From theory to applications},
  author={Ren, Jintao and Bi, Ziqian and Niu, Qian and Song, Xinyuan and Jiang, Zekun and Liu, Junyu and Peng, Benji and Zhang, Sen and Pan, Xuanhe and Wang, Jinlang and others},
  journal={arXiv preprint arXiv:2410.15584},
  year={2024}
}

@inproceedings{chen2024language,
  title={Language-augmented symbolic planner for open-world task planning},
  author={Chen, Guanqi and Yang, Lei and Jia, Ruixing and Hu, Zhe and Chen, Yizhou and Zhang, Wei and Wang, Wenping and Pan, Jia},
  booktitle={Proceedings of Robotics: Science and Systems (RSS)},
  year={2024},
  doi={10.15607/RSS.2024.XX.037}
}

@inproceedings{ozen2015practical,
  title={Practical {PID} controller tuning for motion control},
  author={Ozen, Ozhan and Sariyildiz, Emre and Yu, Haoyong and Ogawa, Kenji and Ohnishi, Kouhei and Sabanovic, Asif},
  booktitle={2015 IEEE International Conference on Mechatronics (ICM)},
  pages={240--245},
  year={2015},
  organization={IEEE}
}

@inproceedings{sui2025grounding,
  title={From grounding to manipulation: Case studies of foundation model integration in embodied robotic systems},
  author={Sui, Xiuchao and Tian, Daiying and Sun, Qi and Chen, Ruirui and Choi, Dongkyu and Kwok, Kenneth and Poria, Soujanya},
  booktitle={Findings of the Association for Computational Linguistics: EMNLP 2025},
  pages={1324--1340},
  year={2025},
  doi={10.18653/v1/2025.findings-emnlp.69}
}

@inproceedings{stone2023open,
  title={Open-world object manipulation using pre-trained vision-language models},
  author={Stone, Austin and Xiao, Ted and Lu, Yao and Gopalakrishnan, Keerthana and Lee, Kuang-Huei and Vuong, Quan and Wohlhart, Paul and Kirmani, Sean and Zitkovich, Brianna and Xia, Fei and others},
  booktitle={Proceedings of the 7th Conference on Robot Learning (CoRL)},
  series={Proceedings of Machine Learning Research},
  volume={229},
  pages={3397--3417},
  year={2023},
  publisher={PMLR}
}

@inproceedings{yang2024binding,
  title={Binding touch to everything: Learning unified multimodal tactile representations},
  author={Yang, Fengyu and Feng, Chao and Chen, Ziyang and Park, Hyoungseob and Wang, Daniel and Dou, Yiming and Zeng, Ziyao and Chen, Xien and Gangopadhyay, Rit and Owens, Andrew and others},
  booktitle={Proceedings of the IEEE/CVF Conference on Computer Vision and Pattern Recognition},
  pages={26340--26353},
  year={2024}
}

@inproceedings{yu2024octopi,
  title={Octopi: Object property reasoning with large tactile-language models},
  author={Yu, Samson and Lin, Kelvin and Xiao, Anxing and Duan, Jiafei and Soh, Harold},
  booktitle={Proceedings of Robotics: Science and Systems (RSS)},
  year={2024},
  doi={10.15607/RSS.2024.XX.066}
}

@article{zhou2022review,
  title={A review of motion planning algorithms for intelligent robots},
  author={Zhou, Chengmin and Huang, Bingding and Fr{\"a}nti, Pasi},
  journal={Journal of Intelligent Manufacturing},
  volume={33},
  number={2},
  pages={387--424},
  year={2022},
  publisher={Springer}
}

@inproceedings{shah2023lm,
  title={{LM-Nav}: Robotic navigation with large pre-trained models of language, vision, and action},
  author={Shah, Dhruv and Osi{\'n}ski, B{\l}a{\.z}ej and Ichter, Brian and Levine, Sergey},
  booktitle={Conference on Robot Learning},
  pages={492--504},
  year={2023},
  organization={PMLR}
}

@inproceedings{driess2023palm,
  title={{PaLM-E}: An embodied multimodal language model},
  author={Driess, Danny and Xia, Fei and Sajjadi, Mehdi S. M. and Lynch, Corey and Chowdhery, Aakanksha and Ichter, Brian and Wahid, Ayzaan and Tompson, Jonathan and Vuong, Quan and Yu, Tianhe and others},
  booktitle={Proceedings of the 40th International Conference on Machine Learning (ICML)},
  series={Proceedings of Machine Learning Research},
  volume={202},
  pages={8469--8488},
  year={2023},
  publisher={PMLR}
}

@inproceedings{dehghani2023scaling,
  title={Scaling vision transformers to 22 billion parameters},
  author={Dehghani, Mostafa and Djolonga, Josip and Mustafa, Basil and Padlewski, Piotr and Heek, Jonathan and Gilmer, Justin and Steiner, Andreas Peter and Caron, Mathilde and Geirhos, Robert and Alabdulmohsin, Ibrahim and others},
  booktitle={International Conference on Machine Learning},
  pages={7480--7512},
  year={2023},
  organization={PMLR}
}

@inproceedings{brohan2022rt,
  title={{RT-1}: Robotics transformer for real-world control at scale},
  author={Brohan, Anthony and Brown, Noah and Carbajal, Justice and Chebotar, Yevgen and Dabis, Joseph and Finn, Chelsea and Gopalakrishnan, Keerthana and Hausman, Karol and Herzog, Alexander and Hsu, Jasmine and others},
  booktitle={Proceedings of Robotics: Science and Systems (RSS)},
  year={2023},
  doi={10.15607/RSS.2023.XIX.025}
}

@inproceedings{brohan2023rt,
  title={{RT-2}: Vision-language-action models transfer web knowledge to robotic control},
  author={Zitkovich, Brianna and Yu, Tianhe and Xu, Sichun and Xu, Peng and Xiao, Ted and Xia, Fei and Wu, Jialin and Wohlhart, Paul and Welker, Stefan and Wahid, Ayzaan and others},
  booktitle={Proceedings of the 7th Conference on Robot Learning (CoRL)},
  series={Proceedings of Machine Learning Research},
  volume={229},
  pages={2165--2183},
  year={2023},
  publisher={PMLR}
}

@inproceedings{burns2024genchip,
  title={{GenCHiP}: Generating robot policy code for high-precision and contact-rich manipulation tasks},
  author={Burns, Kaylee and Jain, Ajinkya and Go, Keegan and Xia, Fei and Stark, Michael and Schaal, Stefan and Hausman, Karol},
  booktitle={2024 IEEE/RSJ International Conference on Intelligent Robots and Systems (IROS)},
  pages={9596--9603},
  year={2024},
  organization={IEEE}
}

@article{ji2025genswarm,
  title={{GenSwarm}: Scalable multi-robot code-policy generation and deployment via language models},
  author={Ji, Wenkang and Chen, Huaben and Chen, Mingyang and Zhu, Guobin and Xu, Lufeng and Gro{\ss}, Roderich and Zhou, Rui and Cao, Ming and Zhao, Shiyu},
  journal={npj Robotics},
  volume={4},
  number={1},
  pages={5},
  year={2026},
  doi={10.1038/s44182-025-00065-w}
}

@article{vemprala2024chatgpt,
  title={{ChatGPT} for robotics: Design principles and model abilities},
  author={Vemprala, Sai H. and Bonatti, Rogerio and Bucker, Arthur and Kapoor, Ashish},
  journal={IEEE Access},
  volume={12},
  pages={55682--55696},
  year={2024},
  doi={10.1109/ACCESS.2024.3387941}
}

@inproceedings{nasiriany2024pivot,
  title={{PIVOT}: Iterative visual prompting elicits actionable knowledge for {VLMs}},
  author={Nasiriany, Soroush and Xia, Fei and Yu, Wenhao and Xiao, Ted and Liang, Jacky and Dasgupta, Ishita and Xie, Annie and Driess, Danny and Wahid, Ayzaan and Xu, Zhuo and others},
  booktitle={Proceedings of the 41st International Conference on Machine Learning (ICML)},
  series={Proceedings of Machine Learning Research},
  volume={235},
  pages={37321--37341},
  year={2024},
  publisher={PMLR}
}

@article{mon2025embodied,
  title={Embodied large language models enable robots to complete complex tasks in unpredictable environments},
  author={Mon-Williams, Ruaridh and Li, Gen and Long, Ran and Du, Wenqian and Lucas, Christopher G.},
  journal={Nature Machine Intelligence},
  volume={7},
  number={4},
  pages={592--601},
  year={2025},
  doi={10.1038/s42256-025-01005-x}
}

@inproceedings{wu2024highlighting,
  title={On the vulnerability of {LLM/VLM}-controlled robotics},
  author={Wu, Xiyang and Chakraborty, Souradip and Xian, Ruiqi and Liang, Jing and Guan, Tianrui and Liu, Fuxiao and Sadler, Brian M. and Manocha, Dinesh and Bedi, Amrit Singh},
  booktitle={Proceedings of the IEEE/RSJ International Conference on Intelligent Robots and Systems (IROS)},
  pages={1914--1921},
  year={2025},
  doi={10.1109/IROS60139.2025.11246863}
}

@article{xu2024hallucination,
  title={Hallucination is inevitable: An innate limitation of large language models},
  author={Xu, Ziwei and Jain, Sanjay and Kankanhalli, Mohan},
  journal={arXiv preprint arXiv:2401.11817},
  year={2024}
}

@inproceedings{cohen2024survey,
  title={A survey of robotic language grounding: Tradeoffs between symbols and embeddings},
  author={Cohen, Vanya and Liu, Jason Xinyu and Mooney, Raymond and Tellex, Stefanie and Watkins, David},
  booktitle={Proceedings of the International Joint Conference on Artificial Intelligence (IJCAI)},
  pages={7999--8009},
  year={2024},
  doi={10.24963/ijcai.2024/885}
}

@inproceedings{rani2006affective,
  title={Affective feedback in closed loop human-robot interaction},
  author={Rani, Pramila and Liu, Changchun and Sarkar, Nilanjan},
  booktitle={Proceedings of the 1st ACM SIGCHI/SIGART Conference on Human-Robot Interaction},
  pages={335--336},
  year={2006}
}

@article{skubis2023humanoid,
  title={Humanoid robots in managerial positions -- Decision-making process and human oversight},
  author={Skubis, Ida and Wodarski, Krzysztof},
  journal={Scientific Papers of Silesian University of Technology. Organization and Management Series},
  number={189},
  pages={573--596},
  year={2023},
  doi={10.29119/1641-3466.2023.189.36}
}

@article{wu2024towards,
  title={Towards audio language modeling---{A}n overview},
  author={Wu, Haibin and Chen, Xuanjun and Lin, Yi-Cheng and Chang, Kai-wei and Chung, Ho-Lam and Liu, Alexander H and Lee, Hung-yi},
  journal={arXiv preprint arXiv:2402.13236},
  year={2024}
}

@inproceedings{xu2024cross,
  title={Cross-modal safety mechanism transfer in large vision-language models},
  author={Xu, Shicheng and Pang, Liang and Zhu, Yunchang and Shen, Huawei and Cheng, Xueqi},
  booktitle={Proceedings of the International Conference on Learning Representations (ICLR)},
  year={2025}
}

@article{jeong2024survey,
  title={A survey of robot intelligence with large language models},
  author={Jeong, Hyeongyo and Lee, Haechan and Kim, Changwon and Shin, Sungtae},
  journal={Applied Sciences},
  volume={14},
  number={19},
  pages={8868},
  year={2024},
  publisher={MDPI}
}

@article{hu2023toward,
  title={Toward general-purpose robots via foundation models: A survey and meta-analysis},
  author={Hu, Yafei and Xie, Quanting and Jain, Vidhi and Francis, Jonathan and Patrikar, Jay and Keetha, Nikhil and Kim, Seungchan and Xie, Yaqi and Zhang, Tianyi and Fang, Hao-Shu and others},
  journal={arXiv preprint arXiv:2312.08782},
  year={2023}
}

@inproceedings{jones2025beyond,
  title={Beyond sight: Finetuning generalist robot policies with heterogeneous sensors via language grounding},
  author={Jones, Joshua and Mees, Oier and Sferrazza, Carmelo and Stachowicz, Kyle and Abbeel, Pieter and Levine, Sergey},
  booktitle={Proceedings of the IEEE International Conference on Robotics and Automation (ICRA)},
  pages={5961--5968},
  year={2025},
  doi={10.1109/ICRA55743.2025.11127987}
}

@article{ravichandran2025safety,
  title={Safety guardrails for {LLM}-enabled robots},
  author={Ravichandran, Zachary and Robey, Alexander and Kumar, Vijay and Pappas, George J. and Hassani, Hamed},
  journal={IEEE Robotics and Automation Letters},
  volume={11},
  number={4},
  pages={4649--4656},
  year={2026},
  doi={10.1109/LRA.2026.3667488}
}

@inproceedings{katara2024gen2sim,
  title={{Gen2Sim}: Scaling up robot learning in simulation with generative models},
  author={Katara, Pushkal and Xian, Zhou and Fragkiadaki, Katerina},
  booktitle={2024 IEEE International Conference on Robotics and Automation (ICRA)},
  pages={6672--6679},
  year={2024},
  organization={IEEE}
}

@article{lin2025proc4gem,
  title={{Proc4Gem}: Foundation models for physical agency through procedural generation},
  author={Lin, Yixin and Humplik, Jan and Huang, Sandy H and Hasenclever, Leonard and Romano, Francesco and Saliceti, Stefano and Zheng, Daniel and Chen, Jose Enrique and Barros, Catarina and Collister, Adrian and others},
  journal={arXiv preprint arXiv:2503.08593},
  year={2025}
}

@inproceedings{yang2024holodeck,
  title={Holodeck: Language guided generation of {3D} embodied {AI} environments},
  author={Yang, Yue and Sun, Fan-Yun and Weihs, Luca and VanderBilt, Eli and Herrasti, Alvaro and Han, Winson and Wu, Jiajun and Haber, Nick and Krishna, Ranjay and Liu, Lingjie and others},
  booktitle={Proceedings of the IEEE/CVF Conference on Computer Vision and Pattern Recognition},
  pages={16227--16237},
  year={2024}
}

@article{afzal2021gzscenic,
  title={{GzScenic}: Automatic scene generation for {Gazebo} simulator},
  author={Afzal, Afsoon and Goues, Claire Le and Timperley, Christopher S},
  journal={arXiv preprint arXiv:2104.08625},
  year={2021}
}

@misc{karavaev_worldcreator,
  author    = {Alex Karavaev},
  title={{World Creator}},
  year      = {2025},
  url       = {https://github.com/AlexKaravaev/world-creator},
  note      = {Accessed: 2025-06-15}
}

@inproceedings{gao2025genmanip,
  title={{GenManip}: {LLM}-driven simulation for generalizable instruction-following manipulation},
  author={Gao, Ning and Chen, Yilun and Yang, Shuai and Chen, Xinyi and Tian, Yang and Li, Hao and Huang, Haifeng and Wang, Hanqing and Wang, Tai and Pang, Jiangmiao},
  booktitle={Proceedings of the Computer Vision and Pattern Recognition Conference},
  pages={12187--12198},
  year={2025}
}

@article{xu2023creative,
  title={Creative robot tool use with large language models},
  author={Xu, Mengdi and Huang, Peide and Yu, Wenhao and Liu, Shiqi and Zhang, Xilun and Niu, Yaru and Zhang, Tingnan and Xia, Fei and Tan, Jie and Zhao, Ding},
  journal={arXiv preprint arXiv:2310.13065},
  year={2023}
}

@inproceedings{wang2023gensim,
  title={{GenSim}: Generating robotic simulation tasks via large language models},
  author={Wang, Lirui and Ling, Yiyang and Yuan, Zhecheng and Shridhar, Mohit and Bao, Chen and Qin, Yuzhe and Wang, Bailin and Xu, Huazhe and Wang, Xiaolong},
  booktitle={Proceedings of the International Conference on Learning Representations (ICLR)},
  year={2024}
}

@article{chen2024roboscript,
  title={{RoboScript}: Code generation for free-form manipulation tasks across real and simulation},
  author={Chen, Junting and Mu, Yao and Yu, Qiaojun and Wei, Tianming and Wu, Silang and Yuan, Zhecheng and Liang, Zhixuan and Yang, Chao and Zhang, Kaipeng and Shao, Wenqi and others},
  journal={arXiv preprint arXiv:2402.14623},
  year={2024}
}

@article{kolve2017ai2,
  title={{AI2-THOR}: An interactive {3D} environment for visual {AI}},
  author={Kolve, Eric and Mottaghi, Roozbeh and Han, Winson and VanderBilt, Eli and Weihs, Luca and Herrasti, Alvaro and Deitke, Matt and Ehsani, Kiana and Gordon, Daniel and Zhu, Yuke and others},
  journal={arXiv preprint arXiv:1712.05474},
  year={2017}
}

@inproceedings{zawalski2024robotic,
  title={Robotic control via embodied chain-of-thought reasoning},
  author={Zawalski, Micha{\l} and Chen, William and Pertsch, Karl and Mees, Oier and Finn, Chelsea and Levine, Sergey},
  booktitle={Proceedings of the 8th Conference on Robot Learning (CoRL)},
  series={Proceedings of Machine Learning Research},
  volume={270},
  pages={3157--3181},
  year={2025},
  publisher={PMLR}
}

@article{chen2023typefly,
  title={{TypeFly}: Flying drones with large language model},
  author={Chen, Guojun and Yu, Xiaojing and Ling, Neiwen and Zhong, Lin},
  journal={arXiv preprint arXiv:2312.14950},
  year={2023}
}

@article{lim2025taking,
  title={Taking flight with dialogue: Enabling natural language control for {PX4}-based drone agent},
  author={Lim, Shoon Kit and Chong, Melissa Jia Ying and Khor, Jing Huey and Ling, Ting Yang},
  journal={arXiv preprint arXiv:2506.07509},
  year={2025}
}

@misc{nvidia_physicalai_spatial_warehouse,
  author    = {NVIDIA},
  title={{Physical AI Spatial Intelligence Warehouse}},
  year      = {2025},
  url       = {https://huggingface.co/datasets/nvidia/PhysicalAI-Spatial-Intelligence-Warehouse},
  note      = {Accessed: 2025-06-15},
  howpublished = {Hugging Face Dataset Repository}
}

@article{naderi2024foundation,
  title={Foundation models for autonomous robots in unstructured environments: Current state-of-the-art, challenges, and future pathways},
  author={Naderi, Hossein and Shojaei, Alireza and Huang, Lifu},
  journal={Eng},
  volume={7},
  number={7},
  pages={358},
  year={2026},
  doi={10.3390/eng7070358}
}

@inproceedings{szot2023large,
  title={Large language models as generalizable policies for embodied tasks},
  author={Szot, Andrew and Schwarzer, Max and Agrawal, Harsh and Mazoure, Bogdan and Metcalf, Rin and Talbott, Walter and Mackraz, Natalie and Hjelm, R Devon and Toshev, Alexander T},
  booktitle={Proceedings of the International Conference on Learning Representations (ICLR)},
  year={2024}
}

@article{hao2025embodied,
  title={Humanoid agent via embodied chain-of-action reasoning with multimodal foundation models for zero-shot loco-manipulation},
  author={Wen, Congcong and Bethala, Geeta Chandra Raju and Hao, Yu and Pudasaini, Niraj and Huang, Hao and Yuan, Shuaihang and Huang, Baoru and Nguyen, Anh and Wang, Mengyu and Tzes, Anthony and others},
  journal={arXiv preprint arXiv:2504.09532},
  year={2025}
}

@article{mei2024quadrupedgpt,
  title={{QuadrupedGPT}: Towards a versatile quadruped agent in open-ended worlds},
  author={Mei, Yuting and Wang, Ye and Zheng, Sipeng and Jin, Qin},
  journal={arXiv preprint arXiv:2406.16578},
  year={2024}
}

@article{ahn2024autort,
  title={{AutoRT}: Embodied foundation models for large scale orchestration of robotic agents},
  author={Ahn, Michael and Dwibedi, Debidatta and Finn, Chelsea and Arenas, Montse Gonzalez and Gopalakrishnan, Keerthana and Hausman, Karol and Ichter, Brian and Irpan, Alex and Joshi, Nikhil and Julian, Ryan and others},
  journal={arXiv preprint arXiv:2401.12963},
  year={2024}
}

@article{bjorck2025gr00t,
  title={{GR00T N1}: An open foundation model for generalist humanoid robots},
  author={Bjorck, Johan and Casta{\~n}eda, Fernando and Cherniadev, Nikita and Da, Xingye and Ding, Runyu and Fan, Linxi and Fang, Yu and Fox, Dieter and Hu, Fengyuan and Huang, Spencer and others},
  journal={arXiv preprint arXiv:2503.14734},
  year={2025}
}

@article{ali2025humanoid,
  title={Humanoid World Models: Open World Foundation Models for Humanoid Robotics},
  author={Ali, Muhammad Qasim and Sridhar, Aditya and Matiana, Shahbuland and Wong, Alex and Al-Sharman, Mohammad},
  journal={arXiv preprint arXiv:2506.01182},
  year={2025}
}

@article{chukwurah2024sim,
  title={Sim-to-real transfer in robotics: Addressing the gap between simulation and real-world performance},
  author={Chukwurah, Naomi and Adebayo, Abiodun Sunday and Ajayi, Olanrewaju Oluwaseun},
  journal={Journal of Frontiers in Multidisciplinary Research},
  volume={5},
  number={1},
  pages={33--39},
  year={2024},
  doi={10.54660/.ijfmr.2024.5.1.33-39}
}

@inproceedings{yu2024natural,
  title={Natural language can help bridge the {Sim2Real} gap},
  author={Yu, Albert and Foote, Adeline and Mooney, Raymond and Mart{\'\i}n-Mart{\'\i}n, Roberto},
  booktitle={Proceedings of Robotics: Science and Systems (RSS)},
  year={2024},
  doi={10.15607/RSS.2024.XX.126}
}

@article{da2025survey,
  title={A survey of sim-to-real methods in {RL}: Progress, prospects and challenges with foundation models},
  author={Da, Longchao and Turnau, Justin and Kutralingam, Thirulogasankar Pranav and Velasquez, Alvaro and Shakarian, Paulo and Wei, Hua},
  journal={arXiv preprint arXiv:2502.13187},
  year={2025}
}

@article{noorani2025abstraction,
  title={From abstraction to reality: {DARPA}'s vision for robust sim-to-real autonomy},
  author={Noorani, Erfaun and Serlin, Zachary and Price, Ben and Velasquez, Alvaro},
  journal={AI Magazine},
  volume={46},
  number={2},
  pages={e70015},
  year={2025},
  doi={10.1002/aaai.70015}
}

@inproceedings{nair2022r3m,
  title={{R3M}: A universal visual representation for robot manipulation},
  author={Nair, Suraj and Rajeswaran, Aravind and Kumar, Vikash and Finn, Chelsea and Gupta, Abhinav},
  booktitle={Proceedings of the 6th Conference on Robot Learning (CoRL)},
  series={Proceedings of Machine Learning Research},
  volume={205},
  pages={892--909},
  year={2023},
  publisher={PMLR}
}

@inproceedings{balazadeh2024synthetic,
  title={Physics context builders: A modular framework for physical reasoning in vision-language models},
  author={Balazadeh, Vahid and Ataei, Mohammadmehdi and Cheong, Hyunmin and Khasahmadi, Amir Hosein and Krishnan, Rahul G.},
  booktitle={Proceedings of the IEEE/CVF International Conference on Computer Vision (ICCV)},
  pages={7318--7328},
  year={2025},
  doi={10.1109/ICCV51701.2025.00687}
}

@inproceedings{liu2025fetchbot,
  title={{FetchBot}: Learning generalizable object fetching in cluttered scenes via zero-shot {Sim2Real}},
  author={Liu, Weiheng and Wan, Yuxuan and Wang, Jilong and Kuang, Yuxuan and Shi, Xuesong and Li, Haoran and Zhao, Dongbin and Zhang, Zhizheng and Wang, He},
  booktitle={Proceedings of the 9th Conference on Robot Learning (CoRL)},
  series={Proceedings of Machine Learning Research},
  volume={305},
  pages={859--884},
  year={2025},
  publisher={PMLR}
}

@inproceedings{zhao2024exploring,
  title={Exploring generative {AI} for {Sim2Real} in driving data synthesis},
  author={Zhao, Haonan and Wang, Yiting and Bashford-Rogers, Thomas and Donzella, Valentina and Debattista, Kurt},
  booktitle={2024 IEEE Intelligent Vehicles Symposium (IV)},
  pages={3071--3077},
  year={2024},
  organization={IEEE}
}

@article{zhang2025generative,
  title={Generative artificial intelligence in robotic manipulation: A survey},
  author={Zhang, Kun and Yun, Peng and Cen, Jun and Cai, Junhao and Zhu, Didi and Yuan, Hangjie and Zhao, Chao and Feng, Tao and Wang, Michael Yu and Chen, Qifeng and others},
  journal={arXiv preprint arXiv:2503.03464},
  year={2025}
}

@inproceedings{chen2021understanding,
  title={Understanding domain randomization for sim-to-real transfer},
  author={Chen, Xiaoyu and Hu, Jiachen and Jin, Chi and Li, Lihong and Wang, Liwei},
  booktitle={Proceedings of the International Conference on Learning Representations (ICLR)},
  year={2022}
}

@inproceedings{muratore2018domain,
  title={Domain randomization for simulation-based policy optimization with transferability assessment},
  author={Muratore, Fabio and Treede, Felix and Gienger, Michael and Peters, Jan},
  booktitle={Conference on Robot Learning},
  pages={700--713},
  year={2018},
  organization={PMLR}
}

@article{yu2025adept,
  title={{ADEPT}: Adaptive diffusion environment for policy transfer sim-to-real},
  author={Yu, Youwei and Xu, Junhong and Liu, Lantao},
  journal={arXiv preprint arXiv:2506.01759},
  year={2025}
}

@article{jang2024bridging,
  title={Bridging the simulation-to-real gap of depth images for deep reinforcement learning},
  author={Jang, Yoonsu and Baek, Jongchan and Jeon, Soo and Han, Soohee},
  journal={Expert Systems with Applications},
  volume={253},
  pages={124310},
  year={2024},
  publisher={Elsevier}
}

@inproceedings{biruduganti2025bridging,
  title={Bridging the {Sim2Real} gap: A framework for evaluating vision encoder pre-training for visuomotor policy transfer},
  author={Yardi, Yash and Biruduganti, Samuel and Ankile, Lars},
  booktitle={Proceedings of the International Conference on Electrical and Computer Engineering Researches (ICECER)},
  pages={1--6},
  year={2025},
  doi={10.1109/ICECER65523.2025.11401367}
}

@inproceedings{chen2024sugar,
  title={{SUGAR}: Pre-training {3D} visual representations for robotics},
  author={Chen, Shizhe and Garcia, Ricardo and Laptev, Ivan and Schmid, Cordelia},
  booktitle={Proceedings of the IEEE/CVF Conference on Computer Vision and Pattern Recognition},
  pages={18049--18060},
  year={2024}
}

@inproceedings{radosavovic2023real,
  title={Real-world robot learning with masked visual pre-training},
  author={Radosavovic, Ilija and Xiao, Tete and James, Stephen and Abbeel, Pieter and Malik, Jitendra and Darrell, Trevor},
  booktitle={Conference on Robot Learning},
  pages={416--426},
  year={2023},
  organization={PMLR}
}

@inproceedings{ma2024dreureka,
  title={{DrEureka}: Language model guided sim-to-real transfer},
  author={Ma, Yecheng Jason and Liang, William and Wang, Hung-Ju and Zhu, Yuke and Fan, Linxi and Bastani, Osbert and Jayaraman, Dinesh},
  booktitle={Proceedings of Robotics: Science and Systems (RSS)},
  year={2024},
  doi={10.15607/RSS.2024.XX.094}
}

@article{chen2024rlingua,
  title={{RLingua}: Improving reinforcement learning sample efficiency in robotic manipulations with large language models},
  author={Chen, Liangliang and Lei, Yutian and Jin, Shiyu and Zhang, Ying and Zhang, Liangjun},
  journal={IEEE Robotics and Automation Letters},
  volume={9},
  number={7},
  pages={6075--6082},
  year={2024},
  doi={10.1109/LRA.2024.3400189}
}

@article{jiao2023swarm,
  title={{Swarm-GPT}: Combining large language models with safe motion planning for robot choreography design},
  author={Jiao, Aoran and Patel, Tanmay P and Khurana, Sanjmi and Korol, Anna-Mariya and Brunke, Lukas and Adajania, Vivek K and Culha, Utku and Zhou, Siqi and Schoellig, Angela P},
  journal={arXiv preprint arXiv:2312.01059},
  year={2023}
}

@inproceedings{an2025rag,
  title={{RAG} {LLMs} are not safer: A safety analysis of retrieval-augmented generation for large language models},
  author={An, Bang and Zhang, Shiyue and Dredze, Mark},
  booktitle={Proceedings of the 2025 Conference of the Nations of the Americas Chapter of the Association for Computational Linguistics: Human Language Technologies (NAACL)},
  pages={5444--5474},
  year={2025},
  publisher={Association for Computational Linguistics},
  doi={10.18653/v1/2025.naacl-long.281}
}

@article{waheed2025quantifying,
  title={Quantifying the {Sim2Real} gap: Model-based verification and validation in autonomous ground systems},
  author={Waheed, Ammar and Areti, Madhu and Gallantree, Luke and Hasnain, Zohaib},
  journal={IEEE Robotics and Automation Letters},
  volume={10},
  number={4},
  pages={3819--3826},
  year={2025},
  doi={10.1109/LRA.2025.3546126}
}

@article{zahedifar2025llm,
  title={{LLM-Controller}: Dynamic robot control adaptation using large language models},
  author={Zahedifar, Rasoul and Baghshah, Mahdieh Soleymani and Taheri, Alireza},
  journal={Robotics and Autonomous Systems},
  volume={186},
  pages={104913},
  year={2025},
  publisher={Elsevier}
}

@inproceedings{ishimizu2024towards,
  title={Towards efficient discrete controller synthesis: Semantics-aware stepwise policy design via {LLM}},
  author={Ishimizu, Yusei and Li, Jialong and Yamauchi, Takuto and Chen, Sinan and Cai, Jinyu and Hirano, Takanori and Tei, Kenji},
  booktitle={2024 IEEE International Conference on Consumer Electronics-Asia (ICCE-Asia)},
  pages={1--4},
  year={2024},
  organization={IEEE}
}

@article{singh2023progprompt,
  title={{ProgPrompt}: Program generation for situated robot task planning using large language models},
  author={Singh, Ishika and Blukis, Valts and Mousavian, Arsalan and Goyal, Ankit and Xu, Danfei and Tremblay, Jonathan and Fox, Dieter and Thomason, Jesse and Garg, Animesh},
  journal={Autonomous Robots},
  volume={47},
  number={8},
  pages={999--1012},
  year={2023},
  publisher={Springer}
}

@article{gerstmayr2024multibody,
  title={Multibody models generated from natural language},
  author={Gerstmayr, Johannes and Manzl, Peter and Pieber, Michael},
  journal={Multibody System Dynamics},
  volume={62},
  number={2},
  pages={249--271},
  year={2024},
  publisher={Springer}
}

@article{sun2024leveraging,
  title={Leveraging large language models for comprehensive locomotion control in humanoid robots design},
  author={Sun, Shilong and Li, Chiyao and Zhao, Zida and Huang, Haodong and Xu, Wenfu},
  journal={Biomimetic Intelligence and Robotics},
  volume={4},
  number={4},
  pages={100187},
  year={2024},
  publisher={Elsevier}
}

@article{skreta2024replan,
  title={{RePLan}: Robotic replanning with perception and language models},
  author={Skreta, Marta and Zhou, Zihan and Yuan, Jia Lin and Darvish, Kourosh and Aspuru-Guzik, Al{\'a}n and Garg, Animesh},
  journal={arXiv preprint arXiv:2401.04157},
  year={2024}
}

@inproceedings{ouyang2024long,
  title={Long-horizon locomotion and manipulation on a quadrupedal robot with large language models},
  author={Ouyang, Yutao and Li, Jinhan and Li, Yunfei and Li, Zhongyu and Yu, Chao and Sreenath, Koushil and Wu, Yi},
  booktitle={Proceedings of the IEEE/RSJ International Conference on Intelligent Robots and Systems (IROS)},
  pages={11157--11164},
  year={2025},
  doi={10.1109/IROS60139.2025.11246632}
}

@article{ajay2023compositional,
  title={Compositional foundation models for hierarchical planning},
  author={Ajay, Anurag and Han, Seungwook and Du, Yilun and Li, Shuang and Gupta, Abhi and Jaakkola, Tommi and Tenenbaum, Josh and Kaelbling, Leslie and Srivastava, Akash and Agrawal, Pulkit},
  journal={Advances in Neural Information Processing Systems},
  volume={36},
  pages={22304--22325},
  year={2023}
}

@article{glocker2025llm,
  title={{LLM}-empowered embodied agent for memory-augmented task planning in household robotics},
  author={Glocker, Marc and H{\"o}nig, Peter and Hirschmanner, Matthias and Vincze, Markus},
  journal={arXiv preprint arXiv:2504.21716},
  year={2025}
}

@article{awais2025foundation,
  title={Foundation models defining a new era in vision: A survey and outlook},
  author={Awais, Muhammad and Naseer, Muzammal and Khan, Salman and Anwer, Rao Muhammad and Cholakkal, Hisham and Shah, Mubarak and Yang, Ming-Hsuan and Khan, Fahad Shahbaz},
  journal={IEEE Transactions on Pattern Analysis and Machine Intelligence},
  volume={47},
  number={4},
  pages={2245--2264},
  year={2025},
  doi={10.1109/TPAMI.2024.3506283}
}

@article{huang2023instruct2act,
  title={{Instruct2Act}: Mapping multi-modality instructions to robotic actions with large language model},
  author={Huang, Siyuan and Jiang, Zhengkai and Dong, Hao and Qiao, Yu and Gao, Peng and Li, Hongsheng},
  journal={arXiv preprint arXiv:2305.11176},
  year={2023}
}

@article{cong2025overview,
  title={An Overview of Robot Embodied Intelligence Based on Multimodal Models: Tasks, Models, and System Schemes},
  author={Cong, Yao and Mo, Hongwei},
  journal={International Journal of Intelligent Systems},
  volume={2025},
  number={1},
  pages={5124400},
  year={2025},
  publisher={Wiley Online Library}
}

@mastersthesis{shenawa2025task,
  title={Task specific evaluation of large language models: A study for human-robot interaction},
  author={Shenawa, Aiman},
  school={KTH Royal Institute of Technology},
  address={Stockholm, Sweden},
  year={2025},
  url={https://www.diva-portal.org/smash/get/diva2:1947373/FULLTEXT02.pdf}
}

@article{team2023gemini,
  title={{Gemini}: A family of highly capable multimodal models},
  author={{Gemini Team} and Anil, Rohan and Borgeaud, Sebastian and Alayrac, Jean-Baptiste and Yu, Jiahui and Soricut, Radu and Schalkwyk, Johan and Dai, Andrew M and Hauth, Anja and Millican, Katie and others},
  journal={arXiv preprint arXiv:2312.11805},
  year={2023}
}

@article{khan2024leveraging,
  title={Leveraging vision-language models for manufacturing feature recognition in {CAD} designs},
  author={Khan, Muhammad Tayyab and Chen, Lequn and Ng, Ye Han and Feng, Wenhe and Tan, Nicholas Yew Jin and Moon, Seung Ki},
  journal={arXiv preprint arXiv:2411.02810},
  year={2024}
}

@article{khan2024fine,
  title={Fine-Tuning Vision-Language Model for Automated Engineering Drawing Information Extraction},
  author={Khan, Muhammad Tayyab and Chen, Lequn and Ng, Ye Han and Feng, Wenhe and Tan, Nicholas Yew Jin and Moon, Seung Ki},
  journal={arXiv preprint arXiv:2411.03707},
  year={2024}
}

\end{document}